\pdfoutput=1
\documentclass[10pt,twocolumn,letterpaper]{article}

\usepackage[pagenumbers]{cvpr} % To force page numbers, e.g. for an arXiv version
\makeatletter
\@namedef{ver@everyshi.sty}{}
\makeatother
\usepackage{tikz}
\usepackage{graphicx}
\usepackage{amsmath}
\usepackage{amssymb}
\usepackage{booktabs}
\usepackage{latexsym}

\usepackage{url}
\usepackage{booktabs}
\usepackage{multirow}
\usepackage{array, caption, floatrow, makecell, booktabs}
\usepackage{wrapfig}
\usepackage{floatrow}
\usepackage{comment}
\usepackage{enumitem}
\usepackage{accents}
\usepackage{xspace,mfirstuc,tabulary}
\usepackage{tabu}

\usepackage{bm}
\usepackage{todonotes}
\makeatletter
\newcommand*\iftodonotes{\if@todonotes@disabled\expandafter\@secondoftwo\else\expandafter\@firstoftwo\fi}  % defines \iftodonotes{<true>}{<false>}, thanks to https://tex.stackexchange.com/questions/126559/conditional-based-on-packageoption
\makeatother
\usepackage[pagebackref,breaklinks,colorlinks]{hyperref}
\usepackage{pifont}% http://ctan.org/pkg/pifont
\newcommand{\cmark}{\ding{51}}%
\newcommand{\xmark}{\ding{55}}%

\usepackage{amsmath}
\usepackage{amsfonts}
\usepackage{amssymb}

\usepackage{wrapfig}

\newcommand{\RN}[1]{%
	\textup{\lowercase\expandafter{\it \romannumeral#1}}%
}

\usepackage{xcolor}
\definecolor{mygreen}{HTML}{3cb44b}
\definecolor{skyblue}{HTML}{beffff}
\definecolor{lightgreen}{HTML}{90ee90}

\newcommand{\beq}{\vspace{0mm}\begin{equation}}
\newcommand{\eeq}{\vspace{0mm}\end{equation}}
\newcommand{\beqs}{\vspace{0mm}\begin{eqnarray}}
\newcommand{\eeqs}{\vspace{0mm}\end{eqnarray}}
\newcommand{\barr}{\begin{array}}
\newcommand{\earr}{\end{array}}

\newcommand{\R}{\mathbb{R}}

\usepackage{caption}

\newcommand{\Lcal}{\mathcal{L}}

\newcommand{\Ccal}{\mathcal{C}}

\newcommand{\Rcal}{\mathcal{R}}

\definecolor{Gray}{gray}{0.93}
\definecolor{Graylight}{gray}{0.95}
\definecolor{Grayheavy}{gray}{0.90}

\usepackage{colortbl}
\definecolor{Gray}{gray}{0.93}

\usepackage{array, caption, floatrow, makecell, booktabs}
\usepackage{wrapfig}
\usepackage{floatrow}
\usepackage{comment}
\usepackage{float}
\usepackage{wrapfig,lipsum,booktabs}

\usepackage[capitalize]{cleveref}
\crefname{section}{Sec.}{Secs.}
\Crefname{section}{Section}{Sections}
\Crefname{table}{Table}{Tables}
\crefname{table}{Tab.}{Tabs.}

\def\cvprPaperID{9280} % *** Enter the CVPR Paper ID here
\def\confName{CVPR}
\def\confYear{2022}

\begin{document}

\title{Grounded Language-Image Pre-training
}

\author{
\textbf{\normalsize{Liunian Harold Li$^{*1\dagger}$, Pengchuan Zhang$^{*2\spadesuit}$, Haotian Zhang$^{*3\dagger}$,
Jianwei Yang$^{2}$, Chunyuan Li$^{2}$, Yiwu Zhong$^{4\dagger}$,}} \\ \textbf{\normalsize{Lijuan Wang$^{5}$, Lu Yuan$^{5}$, Lei Zhang$^{6}$, Jenq-Neng Hwang$^{3}$, Kai-Wei Chang$^{1}$, Jianfeng Gao$^{2}$}}\\
\normalsize{$^1$UCLA, $^2$Microsoft Research, 
$^{3}$University of Washington,} \\
\normalsize{$^4$University of Wisconsin-Madison, $^{5}$Microsoft Cloud
and AI, $^6$International Digital Economy Academy}\\
% \texttt{\scriptsize{ \{liunian.harold.li,kwchang\}@cs.ucla.edu,\{penzhan,jianwei.yang,Chunyuan.Li\}@microsoft.com,}} \\
% \texttt{\scriptsize{\{lijuanw,luyuan,jfgao\}@microsoft.com,\{haotiz,hwang\}@uw.edu},yzhong52@wisc.edu,leizhang@idea.edu.cn}
}

% Liunian Harold Li, Pengchuan Zhang, Haotian Zhang, Jianwei Yang, Chunyuan Li, Yiwu Zhong, Lijuan Wang, Lu Yuan, Lei Zhang, Jenq-Neng Hwang, Jianfeng Gao

\maketitle

\definecolor{Graylight}{gray}{0.95}

\newcommand\blfootnote[1]{%
  \begingroup
  \renewcommand\thefootnote{}\footnote{#1}%
  \addtocounter{footnote}{-1}%
  \endgroup
}
\newcommand{\our}{GLIP\xspace}
\newcommand{\dyground}{GLIP-T (A)\xspace}
\newcommand{\dyheadcoco}{DyHead {{- COCO}}\xspace}
\newcommand{\dyheadobj}{DyHead\xspace}

\newcommand{\objfive}{Object365\xspace}

\newcommand{\oura}{GLIP-T (B)\xspace}
\newcommand{\ourb}{GLIP-T (C)\xspace}
\newcommand{\ourtiny}{GLIP-T\xspace}

\newcommand{\ourd}{GLIP-L\xspace}
\newcommand{\ourlarge}{GLIP-L\xspace}

\newcommand{\objsuffix}{\hspace{7pt} w/ O365\xspace}
\newcommand{\goldgsuffix}{\hspace{7pt} w/ GoldG\xspace}
% dataset names
\newcommand{\goldg}{GoldG\xspace}
\newcommand{\goldgfull}{GoldG+\xspace}

\newcommand{\std}[1]{\tiny{$\pm$#1}}

%%%%%%%%% ABSTRACT
\begin{abstract}
This paper presents a grounded language-image pre-training (GLIP) model for learning \emph{object-level}, \emph{language-aware}, and \emph{semantic-rich} visual representations. GLIP unifies object detection and phrase grounding for pre-training. The unification brings two benefits: 1) it allows GLIP to learn from both detection and grounding data to improve both tasks and bootstrap a good grounding model; 2) GLIP can leverage massive image-text pairs by generating grounding boxes in a self-training fashion, making the learned representations semantic-rich. \blfootnote{$^*$The three authors contributed equally. $^\spadesuit$ Corresponding author.}
\blfootnote{$^\dagger$Work done when interning at Microsoft Research.}In our experiments, we pre-train GLIP on 27M grounding data, including 3M human-annotated and 24M web-crawled image-text pairs.
The learned representations demonstrate strong zero-shot and few-shot transferability to various object-level recognition tasks. 
1) When \emph{directly evaluated} on COCO and LVIS (without seeing any images in COCO during pre-training), GLIP achieves 49.8 AP and 26.9 AP, respectively, surpassing many supervised baselines.\footnote{Supervised baselines on COCO object detection: Faster-RCNN w/ ResNet50 (40.2) or ResNet101 (42.0), and DyHead w/ Swin-Tiny (49.7).} 
2) After \emph{fine-tuned} on COCO, GLIP achieves 60.8 AP on val and 61.5 AP on test-dev, surpassing prior SoTA. 
3) When \emph{transferred} to 13 downstream object detection tasks, 
a 1-shot GLIP rivals with a fully-supervised Dynamic Head. Code is released at \url{https://github.com/microsoft/GLIP}.

% This paper presents a grounded language-image pre-training (GLIP) model for learning object-level, language-aware, and semantic-rich visual representations. GLIP unifies object detection and phrase grounding for pre-training. The unification brings two benefits: 1) it allows GLIP to learn from both detection and grounding data to improve both tasks and bootstrap a good grounding model; 2) GLIP can leverage massive image-text pairs by generating grounding boxes in a self-training fashion, making the learned representation semantic-rich. In our experiments, we pre-train GLIP on 27M grounding data, including 3M human-annotated and 24M web-crawled image-text pairs. The learned representations demonstrate strong zero-shot and few-shot transferability to various object-level recognition tasks. 1) When directly evaluated on COCO and LVIS (without seeing any images in COCO during pre-training), GLIP achieves 49.8 AP and 26.9 AP, respectively, surpassing many supervised baselines. 2) After fine-tuned on COCO, GLIP achieves 60.8 AP on val and 61.5 AP on test-dev, surpassing prior SoTA. 3) When transferred to 13 downstream object detection tasks, a 1-shot GLIP rivals with a fully-supervised Dynamic Head. Code will be released at https://github.com/microsoft/GLIP.

\end{abstract}

%%%%%%%%% BODY TEXT

\section{Introduction}
\label{sec:intro}

Visual recognition models are typically trained to predict a fixed set of pre-determined object categories, which limits their usability in real-world applications since additional labeled data are needed to generalize to new visual concepts and domains. 
CLIP~\cite{radford2021learning} shows that \textit{image-level}
visual representations can be learned effectively on large amounts of raw image-text pairs.
Because the paired texts contain a boarder set of visual concepts than any pre-defined concept pool, the pre-trained CLIP model is so semantically rich that it can be easily transferred to downstream image classification and text-image retrieval tasks in zero-shot settings.
However, to gain fine-grained understanding of images, as required by many tasks,
such as object detection~\cite{ren2015faster, lin2017focal}, segmentation~\cite{long2015fully, chen2017deeplab}, human pose estimation~\cite{xiao2018simple, sun2019deep}, scene understanding~\cite{kuznetsova2018open,Xu_2017,han2021image}, action recognition~\cite{ji2019action}, vision-language understanding \cite{lu2019vilbert,tan2019lxmert,chen2019uniter,su2019vl,li2019visualbert,li2019unicoder,zhou2019unified,li2020oscar,li2020unsupervised,zhang2021vinvl}, 
\emph{object-level} visual representations are highly desired.

\begin{figure}[t]
\centering
\includegraphics[width=\linewidth]{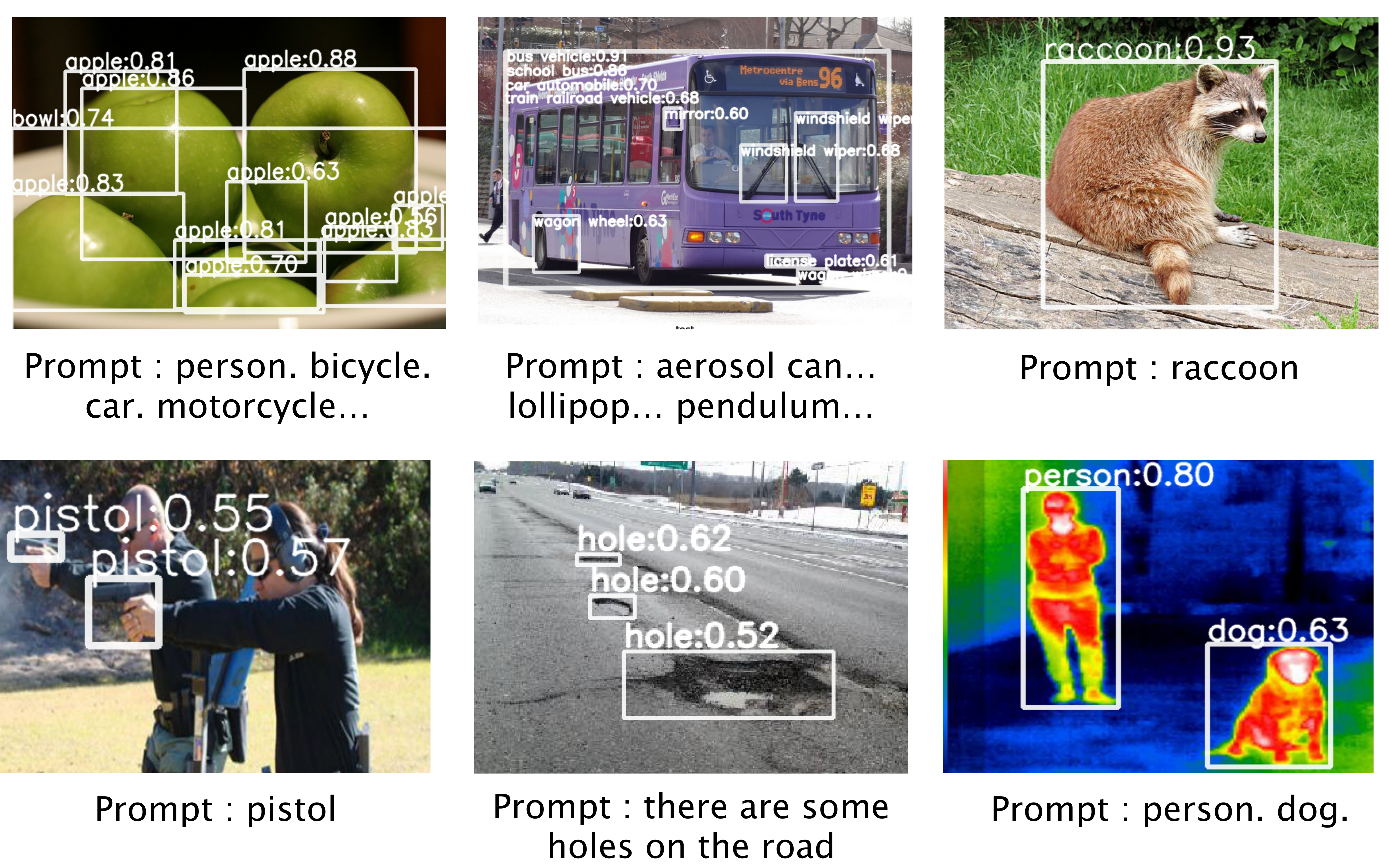}
\caption{GLIP zero-shot transfers to various detection tasks, by writing the categories of interest into a text prompt. }\label{fig:odinw_more_example} % For simplicity, we show only the predictions for the class ``stingray''.
\vspace{-2mm}
\end{figure}

In this paper, we show that \emph{phrase grounding}, which is a task of identifying the fine-grained correspondence between phrases in a sentence and objects (or regions) in an image, is an effective and scalable pre-training task to learn an object-level, language-aware, and semantic-rich visual representation, and propose Grounded Language-Image Pre-training (GLIP).
Our approach unifies the phrase grounding and object detection tasks in that 
%From our perspective, 
object detection can be cast as context-free phrase grounding while phrase grounding can be viewed as a contextualized object detection task. We highlight our key contributions as follows. 

\begin{figure*}[!ht]
\vspace{-4mm}
\centering
\begin{floatrow}
\vspace{-4mm}
  \ffigbox[1.58\FBwidth]{\caption{A unified framework for detection and grounding. Unlike a classical object detection model which predicts a categorical class for each detected object, we reformulate detection as a grounding task by aligning each region/box to phrases in a text prompt. GLIP jointly trains an image encoder and a language encoder to predict the correct pairings of regions and words. We further add the cross-modality deep fusion to early fuse information from two modalities and to learn a language-aware visual representation.}\label{fig:vldyhead}}{%
    \includegraphics[width=\linewidth]{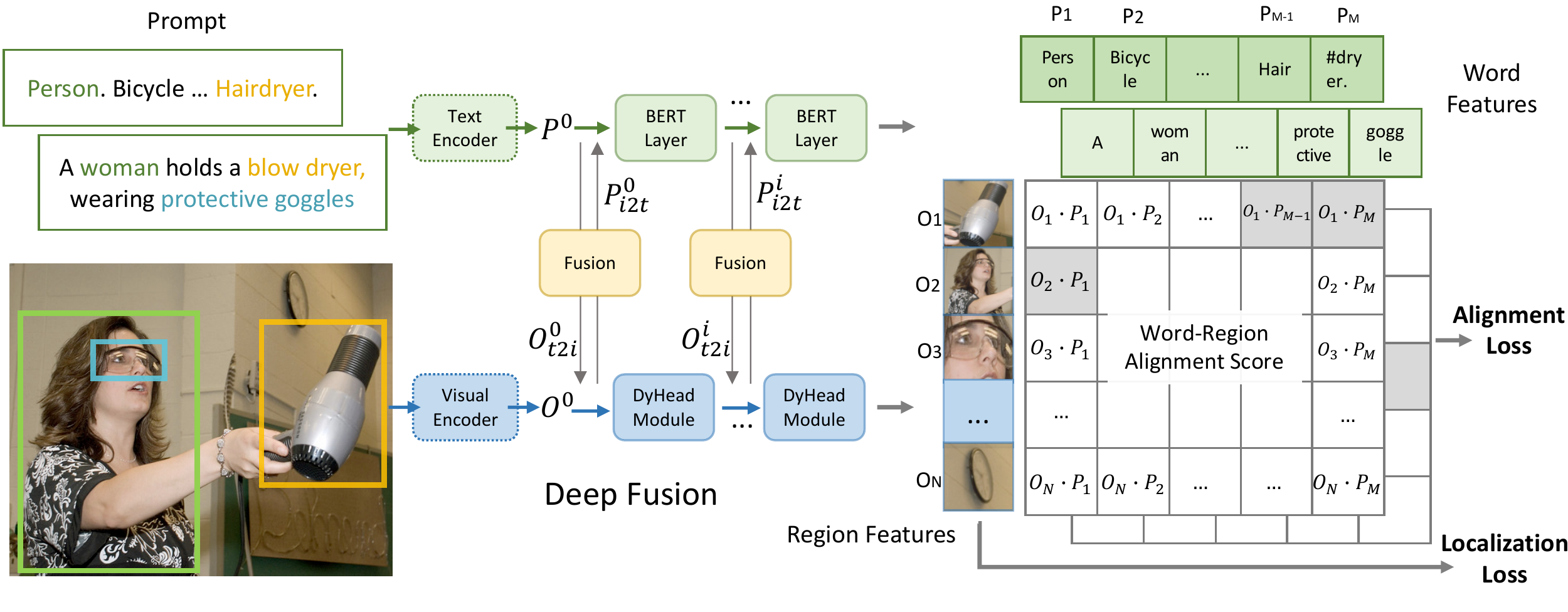} 
  }
  \hspace{-4mm}
  \ffigbox[0.42\FBwidth]{\caption{Grounding predictions from \our. GLIP can locate rare entities, phrases with attributes, and even abstract words.}\label{fig:grounding_example}}{%
    \includegraphics[width=\linewidth]{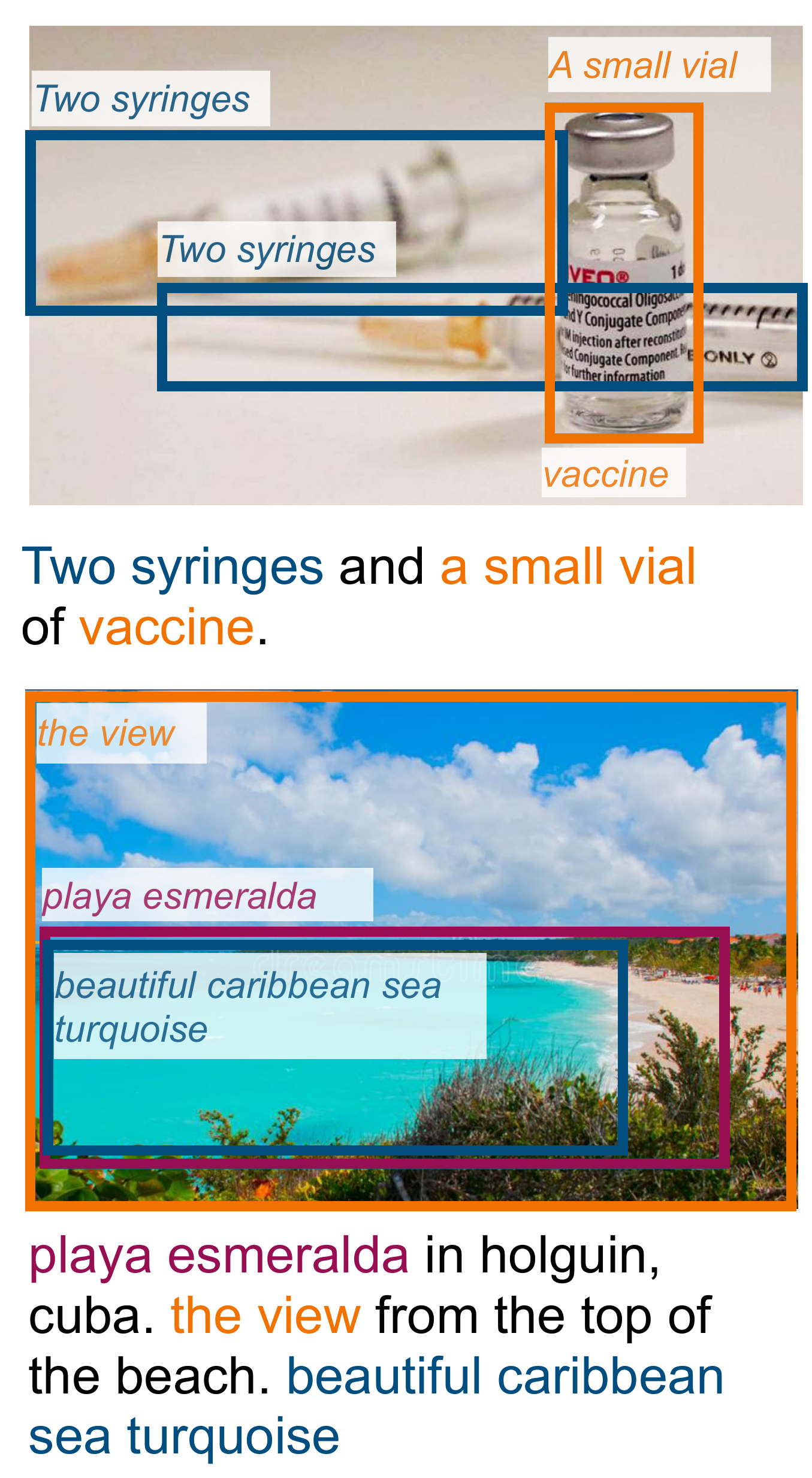}   % Just a dummy. Replace with your figure.
  }
\end{floatrow}
\end{figure*}
{\bf Unifying detection and grounding by reformulating object detection as phrase grounding.}
The reformulation changes the input of a detection model: it takes as input not only an image but also a text prompt that describes \emph{all} the candidate categories in the detection task\footnote{Different from typical phrase grounding tasks, phrases in the text prompt for an object detection task may not be present in the image.}. For example, the text prompt for COCO object detection \cite{lin2014microsoft} is a text string that consists of 80 phrases, i.e., the 80 COCO object class names, joined by ``. '', as shown in Figure~\ref{fig:vldyhead} (Left). 
Any object detection model can be converted to a grounding model by replacing the object classification logits in its box classifier with the word-region alignment scores, i.e., dot product of the region (or box) visual features and the token (or phrase) language features, as shown in Figure~\ref{fig:vldyhead} (Right). 
The language features are computed using a language model, which gives the new detection (or grounding) model a dual-encoder structure. 
Different from CLIP that fuses vision and language only at the last dot product layer~\cite{radford2021learning}, we show that deep cross-modality fusion applied by GLIP, as shown in Figure~\ref{fig:vldyhead} (Middle), is crucial to learn high-quality language-aware visual representations and to achieve superior transfer learning performance. 
The unification of detection and grounding also allows us to pre-train using both types of data and benefits both tasks. On the detection side, the pool of visual concepts is significantly enriched thanks to the grounding data. On the grounding side, detection data introduce more bounding box annotations and help train a new SoTA phrase grounding model.

\textbf{Scaling up visual concepts with massive image-text data.}
Given a good grounding model (teacher), we can augment GLIP pre-training data by automatically generating grounding boxes for massive image-text-paired data, in which noun phrases are detected by an NLP parser~\cite{bird2009natural}. 
Thus, we can pre-train our (student) GLIP-Large model (\ourd) on 27M grounding data, including 3M human-annotated fine-grained data and 24M web-crawled image-text pairs. For the 24M image-text pairs, there are 78.1M high-confidence ($>0.5$) phrase-box pseudo annotations, with 58.4M unique noun phrases. We showcase two real examples of the generated boxes in Figure \ref{fig:grounding_example}. The teacher model can accurately localize some arguably hard concepts, such as \textit{syringes}, \textit{vaccine}, \textit{beautiful caribbean sea turquoise}, and even abstract words (\textit{the view}). Training on such semantic-rich data delivers a semantic-rich student model. In contrast, prior work on scaling detection data simply cannot predict concepts out of the teacher models' pre-defined vocabulary \cite{zoph2020rethinking}. In this study, we show that this simple strategy of scaling up grounding data is empirically effective, bringing large improvements to LVIS and 13 downstream detection tasks, especially on rare categories (Sections \ref{sec:lvis} and \ref{sec:odinw}). When the pre-trained \ourd model is fine-tuned on COCO, it achieves 60.8 AP on COCO 2017val and 61.5 on test-dev, surpassing the current public SoTA models~\cite{dai2021dynamic,xu2021end} that scale up object detection data in various approaches.

% \begin{table}[t]
% \caption{Under prompt tuning, a single GLIP-L model achieves high performance on 14 tasks. We tune only the embedding of each task's prompt while keeping the model weights unchanged. For PascalVOC (2012 Val), we report both AP50:95 and AP50 using COCO evaluation protocol. See Section \ref{sec:onemodel} for details.}
% \label{table:intro_one_model_for_all}
% \begin{center}
% \resizebox{\linewidth}{!}{
% \begin{tabu}{c@{\hskip9pt} 
% c@{\hskip9pt}c@{\hskip9pt}c@{\hskip9pt} 
% c@{\hskip9pt}c@{\hskip9pt}c@{\hskip9pt}
% c}
% \toprule

% COCO & PascalVOC &
% AerialDrone & 
% Aquarium &
% Rabbits &
% EgoHands &
% Mushrooms

% \\
% 58.8 & 72.9/86.7
% & 23.0 
% & 51.8 
% & 72.0 
% & 75.8 
% & 88.1

% \\
% %\cmidrule{1-8}
% \midrule
% Packages &
% Raccoon &
% Shellfish &
% Vehicles &
% Pistols &
% Pothole &
% Thermal 
% \\

% 75.2 
% & 69.5 
% & 73.6 
% & 72.1 
% & 73.7 
% & 53.5 
% & 81.4 

% \\

% \bottomrule
% \end{tabu}
% }
% \end{center}
% \vspace{-4mm}
% \end{table}

\textbf{Transfer learning with GLIP: one model for all.} 
The grounding reformulation and semantic-rich pre-training facilitate domain transfer. GLIP can be transferred to various tasks with few or even no additional human annotations. When the \ourd model is directly evaluated on the COCO and LVIS datasets (without seeing any images in COCO during pre-training), it achieves 49.8 and 26.9 AP on COCO val2017 and LVIS val, respectively, surpassing many supervised baselines. When evaluated on 13 existing object detection datasets, spanning scenarios including fine-grained species detection, drone-view detection, and ego-centric detection, the setting which we term ``Object Detection in the Wild'' (ODinW) (Section \ref{sec:data_effidiency}), GLIP exhibits excellent data efficiency. For example, a zero-shot \ourd outperforms a 10-shot supervised baseline (Dynamic Head) pre-trained on Objects365 while a 1-shot \ourd rivals with a fully supervised Dynamic Head. Moreover, when task-specific annotations are available, instead of tuning the whole model, one could tune only the task-specific prompt embedding, while keeping the model parameters unchanged. Under such a prompt tuning setting (Section \ref{sec:onemodel}), one GLIP model can simultaneously perform well on all downstream tasks %(Table \ref{table:intro_one_model_for_all})
, reducing the fine-tuning and deployment cost.

\section{Related Work}
Standard object detection systems are trained to localize a fixed set of object classes predefined in crowd-labeled datasets, such as COCO~\cite{lin2014microsoft}, OpenImages (OI)~\cite{kuznetsova2018open}, Objects365~\cite{shao2019objects365}, and Visual Genome (VG)~\cite{krishna2017visual}, which contains no more than 2,000 object classes. Such human-annotated data are costly to scale up~\cite{wang2022omni}. GLIP presents an affordable solution by reformulating object detection as a phrase grounding (word-to-region matching) problem, and thus enables the use of grounding and massive image-text-paired data. Though our current implementation is built upon Dynamic Head (DyHead)~\cite{dai2021dynamic}, our unified formulation can be generalized to any object detection systems ~\cite{redmon2016you,lin2017focal,dai2021dynamic,dai2016r,ren2015faster,chen2019hybrid,carion2020end,zhu2020deformable,dai2021dynamic}. % for Grounded Language-Image Pre-training.

Recently, there is a trend to develop vision-and-language approaches to visual recognition problems, where vision models are trained with free-form language supervision. For example, CLIP~\cite{radford2021learning} and ALIGN~\cite{jia2021scaling} perform cross-modal contrastive learning on hundreds or thousands of millions of image-text pairs and can directly perform open-vocabulary image classification. By distilling the knowledge from the CLIP/ALIGN model into a two-stage detector, ViLD~\cite{gu2021zero} is proposed to advance zero-shot object detection. Alternatively, MDETR~\cite{kamath2021mdetr} trains an end-to-end model on existing multi-modal datasets which have explicit alignment between phrases in text and objects in image. Our GLIP inherits the semantic-rich and language-aware property of this line of research, achieves SoTA object detection performance and significantly improves the transferability to downstream detection tasks.

This paper focuses on domain transfer for object detection. The goal is to build one pre-trained model that seamlessly transfers to various tasks and domains, in a zero-shot or few-shot manner. 
Our setting differs from zero-shot detection~\cite{bansal2018zero,rahman2020zero,dhamija2020overlooked,zareian2021open,gu2021zero,Rahman_aaai2020}, where some categories are defined as unseen/rare and not present in the training set. We expect \our to perform well on rare categories (Section \ref{sec:lvis}) but we do not explicitly exclude any categories from our training set, because grounding data are so semantically rich that we expect them to cover many rare categories. This resembles the setting in open-vocabulary object detection~\cite{zareian2021open}, which expects raw image-text data to cover many rare categories. A line of work identifies building a open-world object proposal module that could propose any novel object at test time as the key challenge~\cite{zhu2019zero,Zhu_2020_CVPR,wang2020leads,joseph2021towards,kim2022learning}; GLIP offers a new perspective: the model does not need to propose every possible novel objects from an open set; rather it only needs to propose objects mentioned in the text prompt as the detection branch is conditioned on the prompt.

Beyond performance on rare categories, we also consider the transfer cost in real-world scenarios, i.e., how to achieve the best performance with the least amount of data, training budget, and deployment cost (Section \ref{sec:odinw}). In particular, we show that GLIP supports prompt tuning~\cite{lester2021power}, which matches the performance of full fine-tuning but only tunes a fraction of the model parameters. We also present a novel finding that in object detection, prompt tuning is most effective for a model with deep vision-language fusion such as GLIP, while being far less effective for shallow-fused models. 
This stands in contrast to recent work that investigates prompt tuning only for shallow-fused vision-language models such as CLIP ~\cite{zhou2021learning,gao2021clip,zhang2021tip}.

\section{Grounded Language Image Pre-training}
\label{section:method}

Conceptually, object detection and phrase grounding bear a great similarity. They both seek to localize objects and align them to semantic concepts. This synergy motivates us to cast the classical object detection task into a grounding problem and propose a unified formulation (Sec \ref{sec:formulation}). We further propose to add deep fusion between image and text, making the detection model language-aware and thus a strong grounding model (Sec \ref{sec:unif_arch}). With the reformulation and deep fusion, we can pre-train \our on scalable and semantic-rich grounding data (Sec \ref{sec:pretrain_data}).

\subsection{Unified Formulation}
\label{sec:formulation}

\noindent\textbf{Background: object detection.} A typical detection model feeds an input image into a visual encoder $\text{Enc}_{I}$, with CNN \cite{he2016deep,tan2019efficientnet} or Transformer \cite{liu2021swin,zhang2021multi,yang2021focal} as backbone, and extracts region/box features $O$, as shown in Figure~\ref{fig:vldyhead} (Bottom). Each region/box feature is fed into two prediction heads, \ie, a box classifier $\Ccal$ and a box regressor $\Rcal$, which are trained with the classification loss $\Lcal_{\text{cls}}$ and the localization loss $\Lcal_{\text{loc}}$, respectively:
\begin{equation}\label{eqn:loss}
    \Lcal = \Lcal_{\text{cls}} + \Lcal_{\text{loc}}.
\end{equation}
In two-stage detectors, a separate region proposal network (RPN) with RPN loss $\Lcal_{\text{rpn}}$ is used to distinguish foreground from background and refine anchors. Since $\Lcal_{\text{rpn}}$ does not use semantic information of object classes, we merge it into the localization loss $\Lcal_{\text{loc}}$. In one-stage detectors, localization loss $\Lcal_{\text{loc}}$ may also contain the centerness loss~\cite{tian2019fcos}.

The box classifier $\Ccal$ is typically a simple linear layer, and the classification loss $\Lcal_{\text{cls}}$ can be written as:
\begin{equation}\label{eqn:cls_logits}
    O \!=\! \text{Enc}_{I}(\text{Img}), \   S_{\text{cls}} \!=\! O W^T, \  \Lcal_{\text{cls}} \!=\! loss(S_{\text{cls}}; T).
\end{equation}
Here\footnote{$N$ is the number of region/box features, $d$ is the visual feature hidden dimension, $c$ is the number of object classes, and we ignore the bias in the box classifier for simplicity.}, $O \in \R^{N\times d}$ are the object/region/box features of the input image, $W \in \R^{c\times d}$ is the weight matrix of the box classifier $\Ccal$, $S_{\text{cls}} \in \R^{N\times c}$ are the output classification logits, $T \in \{0,1\}^{N\times c}$ is the target matching between regions and classes computed from the classical many-to-1 matching~\cite{redmon2016you,lin2017focal,dai2016r,ren2015faster} or the bipartite Hungarian match~\cite{carion2020end,zhu2020deformable,dai2021dynamic}. $loss(S; T)$ is typically a cross-entropy loss for two-stage detectors and a focal loss~\cite{lin2017focal} for one-stage detectors.

\noindent\textbf{Object detection as phrase grounding.} 
Instead of classifying each region/box into $c$ classes, we reformulate detection as a grounding task, by grounding/aligning each region to $c$ phrases in a text prompt (see Figure~\ref{fig:vldyhead}). How to design a text prompt for a detection task? Given object classes $[\text{person, bicycle, car, ..., toothbrush}]$, one simple way is 
$$\text{Prompt} = \text{``Detect: person, bicycle, car, ... , toothbrush''},$$ 
in which each class name is a candidate phrase to be grounded. One could design better prompts, by providing more expressive descriptions of these classes and/or by exploiting the preference of a pre-trained language model. For example, when the pre-trained BERT model~\cite{devlin2018bert} is used to initialize our language encoder $\text{Enc}_L$, the prompt ``person. bicycle. car. ... . toothbrush'' works better than the more human-friendly prompt described above. We will discuss the prompt design in Section \ref{sec:onemodel}. 

In a grounding model, we compute the alignment scores $S_{\text{ground}}$ between image regions and words in the prompt:
\begin{equation}\label{eqn:ground_logits}
    O \!=\! \text{Enc}_{I}(\text{Img}),~ P \!=\! \text{Enc}_{L}(\text{Prompt}), ~ S_{\text{ground}} \!=\! O P^{\top},
\end{equation}
where $P \in \R^{M \times d}$ is the contextual word/token features from the language encoder and plays a similar role to the weight matrix $W$ in \eqref{eqn:cls_logits}, as shown in Figure~\ref{fig:vldyhead} (Right). 
The grounding model, consisting of both the image encoder $\text{Enc}_I$ and the language encoder $\text{Enc}_L$, is trained end-to-end by minimizing the loss defined in \eqref{eqn:loss} \& \eqref{eqn:cls_logits}, with a simple replacement of the classification logits $S_{\text{cls}}$ in \eqref{eqn:cls_logits} with the region-word aligment scores $S_{\text{ground}}$ in \eqref{eqn:ground_logits}.

However, in \eqref{eqn:cls_logits}, we now have the logits $S_{\text{ground}} \in \R^{N\times M}$ and the target $T \in \{0,1\}^{N\times c}$. The number  of  (sub)-word  tokens $M$ is always larger than the number of phrases $c$ in the text  prompt due to four reasons: 1) some phrases contain multiple words, e.g., ``traffic light''; 2) some single-word phrases are splitted into multiple (sub)-word  tokens, e.g., ``toothbrush" to ``tooth\#" and ``\#brush''; 3) some are the added tokens, such as ``Detect:'', ``,'', special tokens in language models, and 4) a [NoObj] token is added at the end of the tokenized sequence. When the $loss$ is a (focal) binary sigmoid loss (the $loss$ we use in Section \ref{sec:sec4} \& \ref{sec:odinw}), we expand the original target matrix $T \in \{0,1\}^{N\times c}$ to $T' \in \{0,1\}^{N\times M}$ by making all sub-words positive match if a phrase is a positive match and all added tokens negative match to all image features. With this change, the $loss(S_{\text{ground}}; T')$ remains the same. During inference, we average token probabilities as the phrase probability. \footnote{When the $loss$ is a multi-class cross entropy (CE) loss, following MDETR~\cite{kamath2021mdetr}, all box proposals with no positive match are matched to the [NoObj] token. The $loss(S, T')$ becomes a multi-label multi-class CE loss, and we sum token probabilities as phrase probability during inference.} 

\noindent\textbf{Equivalence between detection and grounding.} 
With the above reformulation, we can convert any detection model into a grounding model, and the two views, i.e., detection and grounding, are theoretically equivalent for both training and inference. We also verify this empirically: the SoTA DyHead detector~\cite{dai2021dynamic} with Swin-Tiny backbone gives the same performance on COCO val2017 before and after our reformulation. Please refer to the appendix for discussions. 
With the  reformulation, a pre-trained phrase grounding model can be directly applied to any object detection task, thanks to the free-form input of the language encoder. This makes it possible to transfer our GLIP model to arbitrary detection tasks in a zero-shot manner. 
% \footnote{The equivalence holds when all candidate categories can fit into one prompt. For certain detection tasks (e.g., Objects365 \cite{shao2019objects365}), in practice, we can split the categories into multiple prompts during training and inference.}

\noindent\textbf{Related work.} Our grounding formulation is inspired by MDETR~\cite{kamath2021mdetr}, and our grounding loss %$loss(S_{\text{ground}}; T')$
shares the same spirit of MDETR's fine-grained contrastive loss. We go further than MDETR by finding an effective approach to reformulate detection as grounding and a simple unified loss for both detection and grounding tasks. Our grounding model also resembles models for zero-shot detection~\cite{bansal2018zero,rahman2020zero,gu2021zero,Zhu_2020_CVPR,Rahman_aaai2020}. The seminal work of Bansal et al. \cite{bansal2018zero} enables a detection model to conduct zero-shot detection, by using the pre-trained Glove word embedding~\cite{pennington2014glove} as the phrase features $P \in \R^{c \times d}$, if written in the form of \eqref{eqn:ground_logits}. Recently, phrase features extracted from pre-trained deep language models are introduced in open-vocabulary detection~\cite{zareian2021open}. GLIP differs from zero-shot detection in that GLIP provides a unified view of detection and grounding, and enables two crucial ingredients, \ie, language-aware deep fusion and scaling up with image-text data, as to be described next.

\subsection{Language-Aware Deep Fusion}
\label{sec:unif_arch}

In \eqref{eqn:ground_logits}, the image and text are encoded by separate encoders and only fused at the end to calculate the alignment scores. We call such models \textit{late-fusion} models. 
In vision-language literature~\cite{lu2019vilbert,tan2019lxmert,chen2019uniter,su2019vl,li2019visualbert,li2019unicoder,zhou2019unified,li2020oscar,kamath2021mdetr}, deep fusion of visual and language features is necessary to learn a performant phrase grounding model. We introduce deep fusion between the image and language encoders, which fuses the image and text information in the last few encoding layers, as shown in Figure~\ref{fig:vldyhead} (Middle). Concretely, when we use DyHead~\cite{dai2021dynamic} as the image encoder and BERT~\cite{devlin2018bert} as the text encoder, the deep-fused encoder is:
\begin{align}
    & \textcolor{red}{ O^i_{\text{t2i}}, P^i_{\text{i2t}} } = \text{X-MHA}(O^i, P^i), \label{eq:xmha} \quad i \in \{0, 1, .., L-1\}\\
    & O^{i+1} = \text{DyHeadModule}(O^i \textcolor{red}{ + O^i_{\text{t2i}}}), \quad O = O^{L} \label{eq:i2t}, \\
    & P^{i+1} = \text{BERTLayer}(P^i \textcolor{red}{ + P^i_{\text{i2t}}}), \quad P = P^{L} \label{eq:t2i},
\end{align}
where $L$ is the number of DyHeadModules in DyHead~\cite{dai2021dynamic}, $\text{BERTLayer}$ is {\it newly-added} BERT Layers on top of the pre-trained BERT, $O^0$ denote the visual features from the vision backbone, and $P^0$ denote the token features from the language backbone (BERT). The cross-modality communication is achieved by the cross-modality multi-head attention module (X-MHA)~\eqref{eq:xmha}, followed by the single modality fusion and updated in \eqref{eq:i2t} \& \eqref{eq:t2i}. Without added context vectors (\textcolor{red}{$O^i_{\text{t2i}}$} for vision modality and \textcolor{red}{$P^i_{\text{i2t}}$} for language modality), the model is reduced to a \textit{late-fusion} model. 

In the cross-modality multi-head attention module (X-MHA)~\eqref{eq:xmha}, each head computes the context vectors of one modality by attending to the other modality:
\begin{align*}
    & O^{(q)} \!=\! O W^{(q,I)}, P^{(q)} \!=\! P W^{(q,L)}, Attn \!=\! O^{(q)} (P^{(q)})^{\top}\!/\! \sqrt{d}, \\
    & P^{(v)} = P W^{(v,L)}, ~~ O_{\text{t2i}} = \text{SoftMax}(Attn) P^{(v)} W^{(out,I)}, \\
    & O^{(v)} = O W^{(v,I)}, ~~ P_{\text{i2t}} = \text{SoftMax}(Attn^{\top}) O^{(v)} W^{(out,L)},
\end{align*}
where $ \{W^{(\text{symbol},I)}, W^{(\text{symbol},L)}: \text{symbol} \in \{q, v, out\} \}$ are trainable parameters and play similar roles to those of query, value, and output linear layers in Multi-Head Self-Attention~\cite{vaswani2017attention}, respectively. 

The deep-fused encoder \eqref{eq:xmha}-\eqref{eq:t2i} brings two benefits. 1) It improves the phrase grounding performance. 2) It makes the learned visual features language-aware, and thus the model's prediction is conditioned on the text prompt. This is crucial to achieve the goal of having one model serve all downstream detection tasks (shown in Section \ref{sec:onemodel}). 

\subsection{Pre-training with Scalable Semantic-Rich Data}

\label{sec:pretrain_data}
Considerable efforts have been devoted to collecting detection data that are rich in semantics and large in quantity. % semantically rich, covering as many as rare and fine-grained categories, and 2) large in quantity, covering diverse images for easy transfer to various visual domains \cite{kuznetsova2018open}.
However, human annotations have been proven costy and limited \cite{kuznetsova2018open,gupta2019lvis}. Prior work seeks to scale up in a self-training fashion \cite{zoph2020rethinking}. They use a teacher (a pre-trained detector) to predict boxes from raw images and generate pseudo detection labels to train a student model. But the generated data are still limited in terms of the size of the concept pool, as the teacher can only predict labels defined in the concept pool, constructed on the existing datasets.
In contrast, our model can be trained on both detection and, more importantly, grounding data. We show that grounding data can provide rich semantics to facilitate localization and can be scaled up in a self-training fashion. 

First, the gold grounding data cover a much larger vocabulary of visual concepts than existing detection data. The largest attempts at scaling up detection vocabulary still cover no more than 2,000 categories \cite{krishna2017visual,gupta2019lvis}.
With grounding data, we expand the vocabulary to cover virtually any concepts that appear in the grounded captions. For example, Flickr30K \cite{plummer2015flickr30k} contains 44,518 unique phrases while VG Caption~\cite{krishna2017visual} contains 110,689 unique phrases, orders of magnitude larger than the vocabulary of detection data. We provide an empirical study in Section \ref{sec:analysis_sec4} to show that 0.8M gold grounding data brings a larger improvement on detecting rare categories than additional 2M detection data. 

Further, instead of scaling up detection data, we show a promising route to obtaining semantically rich data: scaling up grounding data. We use a simple approach inspired by self-training. We first pre-train a \textit{teacher} \our with gold (human-annotated) detection and grounding data. Then we use this teacher model to predict boxes for web-collected image-text data, with noun phrases detected by an NLP parser \cite{bird2009natural}. Finally, a  
\textit{student} model is trained with both the gold data and the generated pseudo grounding data. As shown in Figure \ref{fig:grounding_example}, the teacher is capable of generating accurate boxes for semantically rich entities.

\emph{Why can the student model possibly outperform the teacher model?} While discussions remain active in the self-training literature \cite{zoph2020rethinking}, in the context of visual grounding, we posit that the teacher model is utilizing the language context and language generalization ability to accurately ground concepts that it may not inherently know.
For example, in Figure \ref{fig:grounding_example}, the teacher may not directly recognize certain concepts such as \textit{vaccine} and \textit{turquoise}, if they are not present in gold data. However, the rich language context such as syntactic structures can provide strong guidance for the teacher model to perform an ``educated guess''. The model can localize \textit{vaccine} if it can localize \textit{a small vail}; it can localize \textit{turquoise} if it can find \textit{caribbean sea}. When we train the student model, the ``educated guess'' of the teacher model becomes a ``supervised signal'', enabling the student model to learn the concept of \textit{vaccine} and \textit{turquoise}.

\begin{table}[t]
\caption{A detailed list of GLIP model variants.}
\label{table:models}
\begin{center}
\resizebox{\linewidth}{!}{
\begin{tabu}{l@{\hskip9pt} 
|c| c | c 
c  c
 }
\toprule

\multirow{2}{*}{Model} & 
\multirow{2}{*}{Backbone} & 
\multirow{2}{*}{Deep Fusion} & 
 \multicolumn{3}{c}{Pre-Train Data}   \\
  & 
  &   & 
Detection & Grounding & Caption  \\  
 \midrule
GLIP-T (A) & Swin-T & \xmark
  & Objects365 & - & -  \\

 GLIP-T (B) & Swin-T & \cmark
  & Objects365 & - & - \\
 
% GLIP-T (C) & Swin-T & \xmark
% & Objects365 & GoldG & -  \\

GLIP-T (C) & Swin-T & \cmark
& Objects365 & GoldG & -  \\

\midrule

\ourtiny & Swin-T & \cmark & Objects365 & GoldG & Cap4M  \\
\ourd  & Swin-L & \cmark
  & FourODs & GoldG & Cap24M \\

\bottomrule
\end{tabu}
}
\end{center}
\vspace{-6mm}
\end{table}

\section{Transfer to Established Benchmarks}
\label{sec:sec4}

\begin{table*}[t]
\caption{Zero-shot domain transfer and fine-tuning on COCO. \our, without seeing any images from the COCO dataset, can achieve comparable or superior performance than prior supervised models (e.g. \ourtiny under Zero-Shot v.s. Faster RCNN under Fine-Tune). When fully fine-tuned on COCO, \ourlarge surpasses the SoTA performance.}
\label{table:cocomain}
\begin{center}
\resizebox{0.7\linewidth}{!}{
\begin{tabu}{l@{\hskip9pt} 
c@{\hskip9pt} | c | c 
c  c@{\hskip9pt}c@{\hskip9pt}
 }
\toprule

\multirow{2}{*}{Model} & 
\multirow{2}{*}{Backbone} & 
\multirow{2}{*}{Pre-Train Data}  &
{Zero-Shot}  & Fine-Tune   \\

 & 
 & 
 &
\small{2017val}  & \small{2017val / test-dev}   \\

\midrule
\rowfont{\color{darkgray}}
Faster RCNN & RN50-FPN & -  & - & 40.2 / - \\
\rowfont{\color{darkgray}}
Faster RCNN & RN101-FPN & -  & - & 42.0 / - \\
\rowfont{\color{darkgray}}
DyHead-T \cite{dai2021dynamic} & Swin-T & - & - & 49.7 / - \\
\rowfont{\color{darkgray}}
DyHead-L \cite{dai2021dynamic} & Swin-L & - & - & 58.4 / 58.7 \\
\rowfont{\color{darkgray}}
DyHead-L \cite{dai2021dynamic} & Swin-L  & O365,ImageNet21K  & - & 60.3 / 60.6\\

\rowfont{\color{darkgray}} SoftTeacher \cite{xu2021end} & Swin-L  & O365,SS-COCO & - & 60.7 / 61.3 \\

\midrule
DyHead-T & Swin-T &  O365   & 43.6 & 53.3 / - \\

GLIP-T (A) & Swin-T 
  & O365  & 42.9 & 52.9 / - \\

GLIP-T (B) & Swin-T 
  & O365  & 44.9 & 53.8 / - \\

%GLIP-T (C) & Swin-T 
%   & O365,GoldG  & 41.6 & 52.9 / - \\

GLIP-T (C) & Swin-T
& O365,GoldG & \textbf{46.7} & 55.1 / - \\

\ourtiny & Swin-T   &  O365,GoldG,Cap4M & 46.3 & 54.9 / - \\
\ourtiny & Swin-T   & O365,GoldG,CC3M,SBU & 46.6 & \textbf{55.2} / - \\

\midrule

\ourd  & Swin-L  
  & FourODs,GoldG,Cap24M & \textbf{49.8} & \textbf{60.8} / 61.0 \\

\ourd & Swin-L  
  & FourODs,GoldG+,COCO & - & - / \textbf{61.5} \\

\bottomrule
\end{tabu}
}
\end{center}
\vspace{-4mm}
\end{table*}

After pre-training, \our can be applied to grounding and detection tasks with ease. We show strong direct domain transfer performance on three established benchmarks:
1) MS-COCO object detection (COCO) \cite{lin2014microsoft} containing 80 common object categories; 2) LVIS \cite{gupta2019lvis} covering over 1000 objects categories; 3) Flickr30K \cite{plummer2015flickr30k}, for phrase grounding.
We train 5 variants of GLIP (Table \ref{table:models}) to ablate its three core techniques: 1) unified grounding loss; 2) language-aware deep fusion; 3) and pre-training with both types of data. 
Implementation deails are in the appendix.

\noindent\textbf{GLIP-T (A)} is based on a SoTA detection model, Dynamic Head \cite{dai2021dynamic}, with our word-region alignment loss replacing the classification loss. It is based on the Swin-Tiny backbone and pre-trained on O365 (Objects365 \cite{shao2019objects365}), which contains 0.66M images and 365 categories.
As discussed in Section \ref{sec:formulation}, the model can be viewed as a strong classical zero-shot detection model \cite{bansal2018zero}, relying purely on the language encoder to generalize to new concepts.

\noindent\textbf{GLIP-T (B)} is enhanced with language-aware deep fusion but pre-trained only on O365.

\noindent\textbf{GLIP-T (C)} is pre-trained on 1) O365 and 2) GoldG, 0.8M human-annotated gold grounding data curated by MDETR \cite{kamath2021mdetr}, including Flickr30K, VG Caption \cite{krishna2017visual}, and GQA \cite{hudson2019gqa}. We have removed COCO images from the dataset. It is designed to verify the effectiveness of gold grounding data

\noindent\textbf{GLIP-T} is based on the Swin-Tiny backbone and pre-trained on the following data: 1) O365, 2) GoldG as in GLIP-T (C), and 3) Cap4M, 4M image-text pairs collected from the web with boxes generated by GLIP-T (C). We also experiment with existing image caption datasets: CC (Conceptual Captions with 3M data) \cite{sharma2018conceptual} and SBU (with 1M data) \cite{ordonez2011im2text}. We find that CC+SBU \ourtiny performs slightly better than Cap4M \ourtiny on COCO, but slightly worse on the other datasets. For simplicity, we report both versions on COCO but only the Cap4M model for the other tasks. We present the full results in the appendix.

\noindent\textbf{\ourlarge} is based on Swin-Large and trained with: 1) FourODs (2.66M data), 4 detection datasets including Objects365, OpenImages \cite{krasin2017openimages}, Visual Genome (excluding COCO images) \cite{krishna2017visual}, and ImageNetBoxes \cite{krizhevsky2012imagenet}; 2) GoldG as in GLIP-T (C); and 3) CC12M+SBU, 24M image-text data collected from the web with generated boxes.
\begin{table*}[t]
\caption{Zero-shot domain transfer to LVIS. 
While using no LVIS data, GLIP-T/L outperforms strong supervised baselines (shown in gray). Grounding data (both gold and self-supervised) bring large improvements on APr.}
\label{table:zslvis}
\begin{center}
\resizebox{0.7\linewidth}{!}{
\begin{tabu}{l@{\hskip9pt} 
c@{\hskip9pt} | c@{\hskip9pt}c@{\hskip9pt} 
c@{\hskip9pt} c@{\hskip9pt}|c@{\hskip9pt}
c@{\hskip9pt}c@{\hskip9pt}c}
\toprule

\multirow{2}{*}{Model} & 
\multirow{2}{*}{Backbone} 
& \multicolumn{4}{c|}{MiniVal \cite{kamath2021mdetr}} & \multicolumn{4}{c}{Val v1.0}\\

 & & APr & APc & APf & AP & APr & APc & APf & AP \\
\midrule
\rowfont{\color{darkgray}}
 MDETR \cite{kamath2021mdetr} &  RN101  
&  20.9 &  24.9 &  24.3 &  24.2 
&  - &  - &  - &  -  \\
\rowfont{\color{darkgray}}
 MaskRCNN \cite{kamath2021mdetr} &  RN101 &  26.3 &  34.0 &  33.9 &  33.3 
&  - &  - &  - &  - \\
\rowfont{\color{darkgray}}
 Supervised-RFS \cite{gupta2019lvis} & RN50  &  - &  - &  - &  - 
&  12.3 &  24.3 &  32.4 &  25.4 \\

\midrule

GLIP-T (A) & Swin-T  
    & 14.2 & 13.9 & 23.4 & 18.5 
     & 6.0 & 8.0 & 19.4 & 12.3
    \\
GLIP-T (B) & Swin-T
  & 13.5 & 12.8 & 22.2 & 17.8 
  & 4.2 & 7.6 & 18.6 & 11.3 \\
  
%GLIP-T (C) & Swin-T & 15.8 & \textbf{23.0} & 30.8 & \textbf{26.1} & 9.2 & \textbf{15.2} & \textbf{26.5} & \textbf{18.6}
 
GLIP-T (C) & Swin-T & 17.7 & 19.5 & \textbf{31.0} & 24.9
  & 7.5 & 11.6 & \textbf{26.1} & 16.5  \\

\ourtiny & Swin-T
  & \textbf{20.8} & \textbf{21.4} & \textbf{31.0} & \textbf{26.0} 
& \textbf{10.1} & \textbf{12.5} & 25.5 & \textbf{17.2}
\\

\midrule
\multirow{1}{*}{\ourd}  & Swin-L & 
   \textbf{28.2} & \textbf{34.3} & \textbf{41.5} & \textbf{37.3} & \textbf{17.1} & \textbf{23.3} & \textbf{35.4} & \textbf{26.9} \\

\bottomrule
\end{tabu}
}
\end{center}
\vspace{-4mm}
\end{table*}

\begin{table*}[t]
\begin{center}
\resizebox{0.75\linewidth}{!}{
\begin{tabular}{l@{\hskip9pt} 
c@{\hskip9pt} | c@{\hskip9pt} | c@{\hskip9pt}
c@{\hskip9pt} c@{\hskip9pt} |c@{\hskip9pt}
cc@{\hskip9pt}c}
\toprule
\multirow{2}{*}{Row} &
\multirow{2}{*}{Model} & 
\multirow{2}{*}{Data} &
\multicolumn{3}{c|}{Val} &
\multicolumn{3}{c}{Test} \\

& &  & R@1 & R@5 & R@10 & R@1 & R@5 & R@10    \\

\midrule
1 & MDETR-RN101  & GoldG+  & 82.5 & 92.9 & 94.9 & 83.4 & 93.5 & 95.3\\
2 & MDETR-ENB5  & GoldG+ & 83.6 & 93.4 & 95.1 & 84.3 & 93.9 & 95.8 \\

\midrule
3 &  \multirow{3}{*}{\ourtiny}  & GoldG & 84.0 & 95.1 & 96.8 & 84.4 & 95.3 & 97.0\\
  
4 &  & O365,GoldG & 84.8 & 94.9 & 96.3 & 85.5 & 95.4 & 96.6\\
  
5 & & O365,GoldG,Cap4M  & \textbf{85.7} & \textbf{95.4} & \textbf{96.9} & \textbf{85.7} & \textbf{95.8} & \textbf{97.2} \\

\midrule
6 & \multirow{1}{*}{\ourd} 
  
  & FourODs,GoldG,Cap24M  & \textbf{86.7} & \textbf{96.4} & \textbf{97.9}  & \textbf{87.1} & \textbf{96.9} & \textbf{98.1}\\

\bottomrule
\end{tabular}
}
\end{center}
\vspace{-1pt}
\caption{Phrase grounding performance on Flickr30K entities. \ourlarge outperforms previous SoTA by 2.8 points on test R@1.} 
\label{table:flickr}
\vspace{-4mm}
\end{table*}

\subsection{Zero-Shot and Supervised Transfer on COCO}

We conduct experiments on MS-COCO to evaluate models' transfer ability to common categories. We evaluate under two settings: 1) zero-shot domain transfer, and 2) supervised transfer, where we fine-tune the pre-trained models using the standard setting. 
For the fine-tuning setting, we additionally test the performance of a GLIP-L model, where we include the COCO images in the pre-training data (the last row). Specifically, we add the full GoldG+ grounding data and COCO train2017 to the pre-training data. Note that part of COCO 2017val images are present in GoldG+ \cite{kamath2021mdetr}. Thus we only report the test-dev performance of this model. Please see more details in the appendix.

We introduce an additional baseline: DyHead pre-trained on Objects365. We find that COCO 80 categories are fully covered in Objects365. Thus we can evaluate DyHead trained on Objects365 in a ``zero-shot'' way: during inference, instead of predicting from 365 classes, we restrict the model to predict only from the COCO 80 classes.
We list standard COCO detection models for reference. We also list two state-of-the-art models pre-trained with extra data.

Results are present in Table \ref{table:cocomain}. Overall, \our models achieve strong zero-shot and supervised performance. \textbf{Zero-shot GLIP models rival or surpass well-established supervised models}. The best GLIP-T achieves 46.7 AP, surpassing Faster RCNN; \ourd achieves 49.8 AP, surpassing DyHead-T. 
Under the supervised setting, the best \ourtiny brings 5.5 AP improvement upon the standard DyHead (55.2 v.s. 49.7). With the Swin-Large backbone, \textbf{\ourd surpasses the current SoTA on COCO}, reaching 60.8 on 2017val and 61.5 on test-dev, without some bells and whistles in prior SoTA \cite{xu2021end} such as model EMA, mix-up, label smoothing, or soft-NMS. 

We analyze the zero-shot performance of \our and find three contributing factors: close domain overlap between Objects365 and COCO, deep fusion, and grounding data. As Objects365 covers all categories in COCO, the O365 pre-trained DyHead-T shows strong performance, reaching 43.6 zero-shot AP; reformulating the model into a grounding model, we observe a slight performance drop (GLIP-T (A)); adding deep fusion boosts the performance by 2 AP (GLIP-T (B)); the largest contributor is the gold grounding data, with which GLIP-T (C) reaches a zero-shot AP of 46.7. While the addition of image-text data brings slight or no improvement on COCO (GLIP-T v.s. GLIP-T (C)), we find it essential in generalizing to rare classes, as we show in the LVIS experiments.

% We analyze the zero-shot performance of \our and find three contributing factors: close domain overlap between Objects365 and COCO, deep fusion, and grounding data. As Objects365 covers all categories in COCO, the O365 pre-trained DyHead-T shows strong performance, reaching 43.6 zero-shot AP; reformulating the model into a grounding model, we observe a slight performance drop (GLIP-T (A)); adding deep fusion boosts the performance by 2 AP (GLIP-T (B)); the largest contributor is the gold grounding data, with which GLIP-T (C) reaches a zero-shot AP of 46.7. While the addition of image-text data brings slight or no improvement on COCO (GLIP-T v.s. GLIP-T (C)), we find it essential in generalizing to rare classes, as we show in the LVIS experiments.

\subsection{Zero-Shot Transfer on LVIS}
\label{sec:lvis}

We evaluate the model's ability to recognize diverse and rare objects on LVIS in a zero-shot setting.
We report on MiniVal containing 5,000 images introduced in MDETR as well as the full validation set v1.0. Please see the evaluation details in the appendix.  

Results are present in Table \ref{table:zslvis}. We list three supervised models trained on the annotated data of LVIS. \our exhibits strong zero-shot performance on all the categories. \textbf{\ourtiny is on par with supervised MDETR while \ourlarge outperforms Supervised-RFS by a large margin.}

The benefit of using grounding data is evident. Gold grounding data brings a 4.2-point improvement on MiniVal APr (model C  v.s. model B). Adding image-text data further improves performance by 3.1 points. We conclude that the semantic richness of grounding data significantly helps the model recognize rare objects.

\subsection{Phrase Grounding on Flickr30K Entities}
We evaluate the model's ability to ground entities in natural language on Flickr30K entities \cite{plummer2015flickr30k}. Flickr30K is included in the gold grounding data so we directly evaluate the models after pre-training as in MDETR \cite{kamath2021mdetr}. We use the any-box-protocol specified in MDETR. Results are present in Table \ref{table:flickr}. We evaluate three versions of GLIP with different pre-training data. We list the performance of MDETR, the SoTA grounding model. MDETR is trained on GoldG+, containing 1.3M data (GoldG is a subset of GoldG+ excluding COCO images). 

GLIP-T with GoldG (Row 3) achieves similar performance to MDETR with GoldG+, presumably due to the introduction of Swin Transformer, DyHead module, and deep fusion. More interestingly, the addition of detection data helps grounding (Row 4 v.s. 3), showing again the synergy between the two tasks and the effectiveness of our unified loss. Image-text data also helps (Row 5 v.s. 4). Lastly, scaling up (\ourlarge) can achieve 87.1 Recall@1, outperforming the previous SoTA by 2.8 points.

\subsection{Analysis}
\label{sec:analysis_sec4}
In this section, we perform ablation study by pre-training \ourtiny on different data sources (Table \ref{table:data_ablation}). We answer two research questions. 
First, our approach assumes the use of a detection dataset to bootstraps the model. One natural question is whether grounding data brings improvement when paired with different detection data. We find that adding grounding data brings consistent improvement with different detection data (Row 1-6). 

\begin{table}[t]
\caption{Effect of different detection data.}
\label{table:data_ablation}
\begin{center}
\resizebox{\linewidth}{!}{
\begin{tabular}{ 
cc@{\hskip9pt}| c@{\hskip9pt}| c@{\hskip9pt} 
c@{\hskip9pt}c@{\hskip9pt}c@{\hskip9pt}
c@{\hskip9pt}c@{\hskip9pt}c@{\hskip9pt}}
\toprule

\multirow{2}{*}{Row} &
\multirow{2}{*}{Pre-Training Data}  &
 \multicolumn{1}{c|}{COCO} &  & \multicolumn{4}{c}{LVIS MiniVal}\\

  & & 2017val & & AP$_r$ & AP$_c$ & AP$_f$ & AP\\

\midrule

1 & VG w/o COCO & 26.9 && 4.9 & 10.4 & 23.2 & 16.1 \\
2 & + GoldG & 29.2 && 7.8 & 14.0 & 24.5 & 18.5 \\

\midrule
3 & OpenImages  & 29.9 && 12.8 & 12.1 & 17.8 & 14.9 \\
4 & + GoldG & 33.6 && 15.2 & 16.9 & 24.5 & 20.4 \\

\midrule
5 & O365 &  44.9 && 13.5 & 12.8 & 22.2 & 17.8 \\
 6 &+GoldG  & \textbf{46.7} &
  &  17.7 & 19.5 & 31.0 & 24.9 \\
\midrule

7 & O365,GoldG,Cap4M & 46.3 && \textbf{20.8} & 21.4 & 31.0 &  26.0
\\

\midrule
8 & FourODs  & 46.3 &  & 15.0 & \textbf{22.5} & \textbf{32.8} & \textbf{26.8} \\

\bottomrule
\end{tabular}
}
\end{center}
\vspace{-5mm}
\end{table}

% \begin{table}[t]
% \caption{13 ODinW dataset statistics. We summarize the objects of interest for each dataset and report the image number of each split. URLs of the tasks are provided in the appendix. }
% \label{table:odinw_dataset_small}
% \begin{center}
% \resizebox{\linewidth}{!}{
% \begin{tabular}{l@{\hskip9pt} | 
% c@{\hskip9pt}|c}
% \toprule

% Dataset & Objects of Interest & Train/Val/Test  \\
% \midrule
% PascalVOC & Common objects (PascalVOC 2012) & 13690/3422/-  \\
% AerialDrone & Boats, cars, etc. from drone images & 52/15/7 \\
% Aquarium & Penguins, starfish, etc. in an aquarium & 448/127/63 \\
% Rabbits & Cottontail rabbits & 1980/19/10  \\
% EgoHands & Hands in ego-centric images & 3840/480/480 \\
% Mushrooms & Two kinds of mushrooms & 41/5/5\\
% Packages & Delivery packages & 19/4/3 \\

% Raccoon & Raccoon & 150/29/17 \\

% Shellfish & Shrimp, lobster, and crab & 406/116/58 \\

% Vehicles & Car, bus, motorcycle, truck, and ambulance & 878/250/126\\

% Pistols & Pistol & 2377/297/297\\

% Pothole & Potholes on the road & 465/133/67\\

% Thermal & Dogs and people in thermal images & 142/41/20\\

% \bottomrule
% \end{tabular}
% }
% \end{center}
% \end{table}

Second, we have shown the effectiveness of grounding data for both common and rare categories. One orthogonal direction is to scale up detection data by including more images and categories (Section \ref{sec:pretrain_data}). We intend to provide an empirical comparison between scaling up detection data and grounding data. We present \our trained with 4 public detection datasets (Row 8) as an extreme attempt at scaling up detection data with human annotations. The model is trained with 2.66M detection data in total, with an aligned vocabulary of over 1,500 categories. However, it still trails behind Row 6 on COCO and AP$_r$ of LVIS, where Row 6 is trained with only 0.66M detection data and 0.8M gold grounding data. Adding image-text data further widens the gap on LVIS AP$_r$ (20.8 versus 15.0). We conclude that grounding data are indeed more semantic-rich and a promising alternative to scaling up detection data.

\section{Object Detection in the Wild}
\label{sec:odinw}
To evaluate GLIP's transferability to diverse real-world tasks, we curate an ``Object Detection in the Wild'' (ODinW) setting. We choose 13 public datasets on Roboflow\footnote{\url{https://public.roboflow.com/object-detection}}, each requiring a different localization skill. Many of the datasets are designed with a specific application purpose to mimic real-world deployment scenarios. For example, EgoHands requires locating hands of a person; Pothole concerns detecting holes on the road; ThermalDogsandPeople involves identifying dogs and persons in infrared images. Please refer to the appendix for details.

We demonstrate that GLIP facilitates transfer to such diverse tasks. (1) GLIP brings great data efficiency, reaching the same performance with significantly less task-specific data than baselines (Section \ref{sec:data_effidiency}). (2) GLIP enables new domain transfer strategies: when adapting to a new task, we can simply change the text prompt and keep the entire grounding model unchanged. This greatly reduces deployment cost because it allows one centralized model to serve various downstream tasks (Section \ref{sec:onemodel}).

\begin{figure}[t]
\centering
\includegraphics[width=0.85\linewidth]{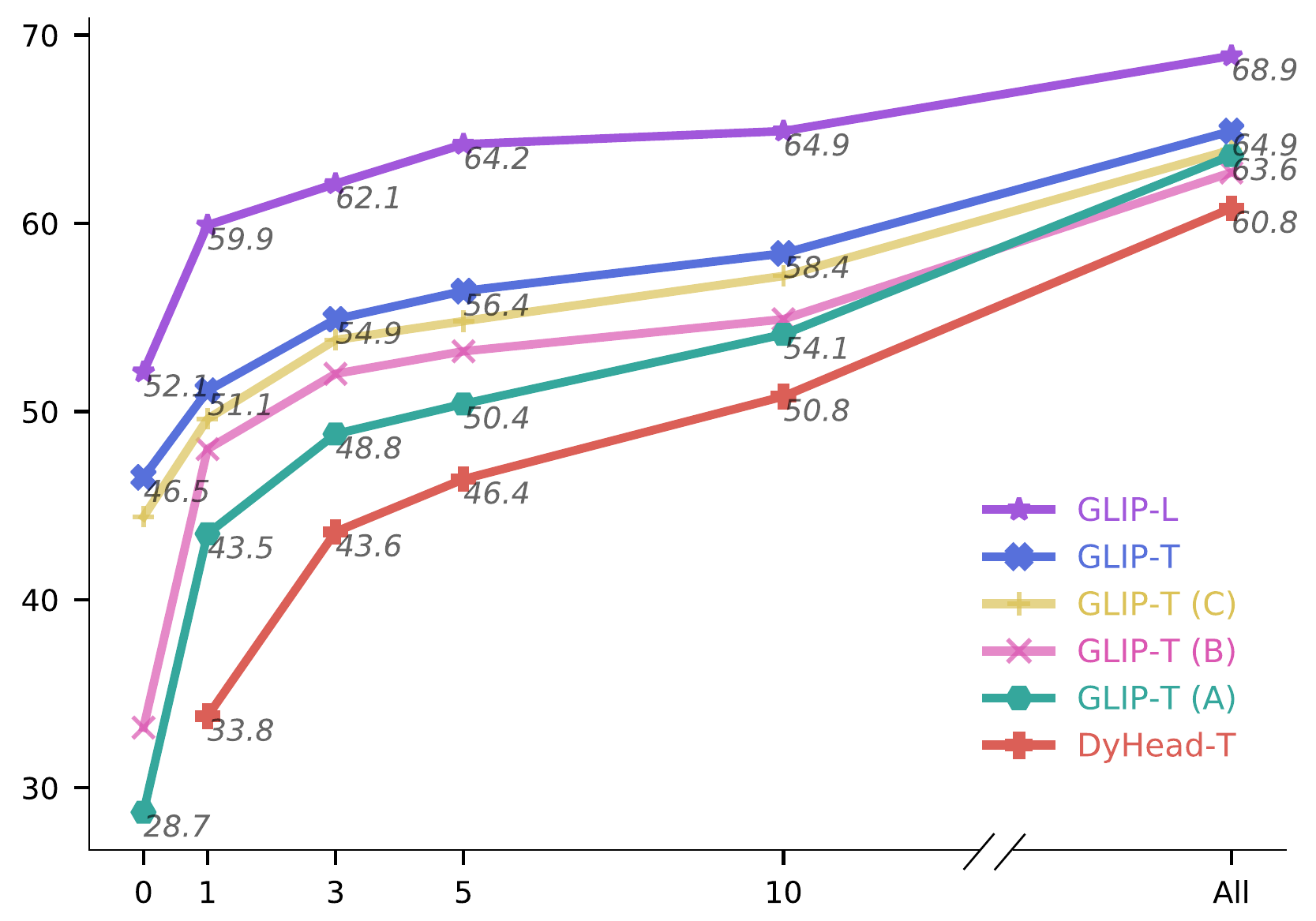}
\caption{Data efficiency of models. X-axis is the amount of task-specific data, from zero-shot to all data. Y-axis is the average AP across 13 datasets. \our exhibits great data efficiency, while each of our proposed approach contributes to the data efficiency.}
\label{fig:data_eff}
\vspace{-3mm}
\end{figure}

\subsection{Data Efficiency}
\label{sec:data_effidiency}
We vary the amount of task-specific annotated data, from zero-shot (no data provided), to $X$-shot (providing at least $X$ examples per category \cite{kang2019few,yan2019meta,wang2019meta}), to using all data in the training set. We fine-tune the models on the provided data and use the same hyper-parameters for all models. Each dataset comes with pre-specified category names. As \our is language-aware, we find it beneficial to re-write some pre-specified names with more descriptive language (see Section \ref{sec:onemodel} for a discussion). We compare with the SoTA detector DyHead-T, pre-trained on Objects365. We test with the standard COCO-trained DyHead-T and find it giving similar performance. For simplicity, we report only the former. We also experiment with the scaled cosine similarity approach \cite{wang2020frustratingly} but find it slightly underperforming the vanilla approach so we report only the latter.
Please refer to the appendix for full statistics, including three independent runs for $X$-shot experiments. 

Results are shown in Figure \ref{fig:data_eff}. We find that unified grounding reformulation, deep fusion, grounding data, and model scale-up all contribute to the improved data efficiency (from the bottom red line (Dyhead-T) up to the upper purple line (GLIP-L)). As a result, \our exhibits transformative data efficiency. A zero-shot \ourtiny outperforms 5-shot DyHead-T while a one-shot \ourd is competitive with a fully supervised DyHead-T. 

\begin{figure} [t]
\centering
\includegraphics[width=0.8\textwidth]{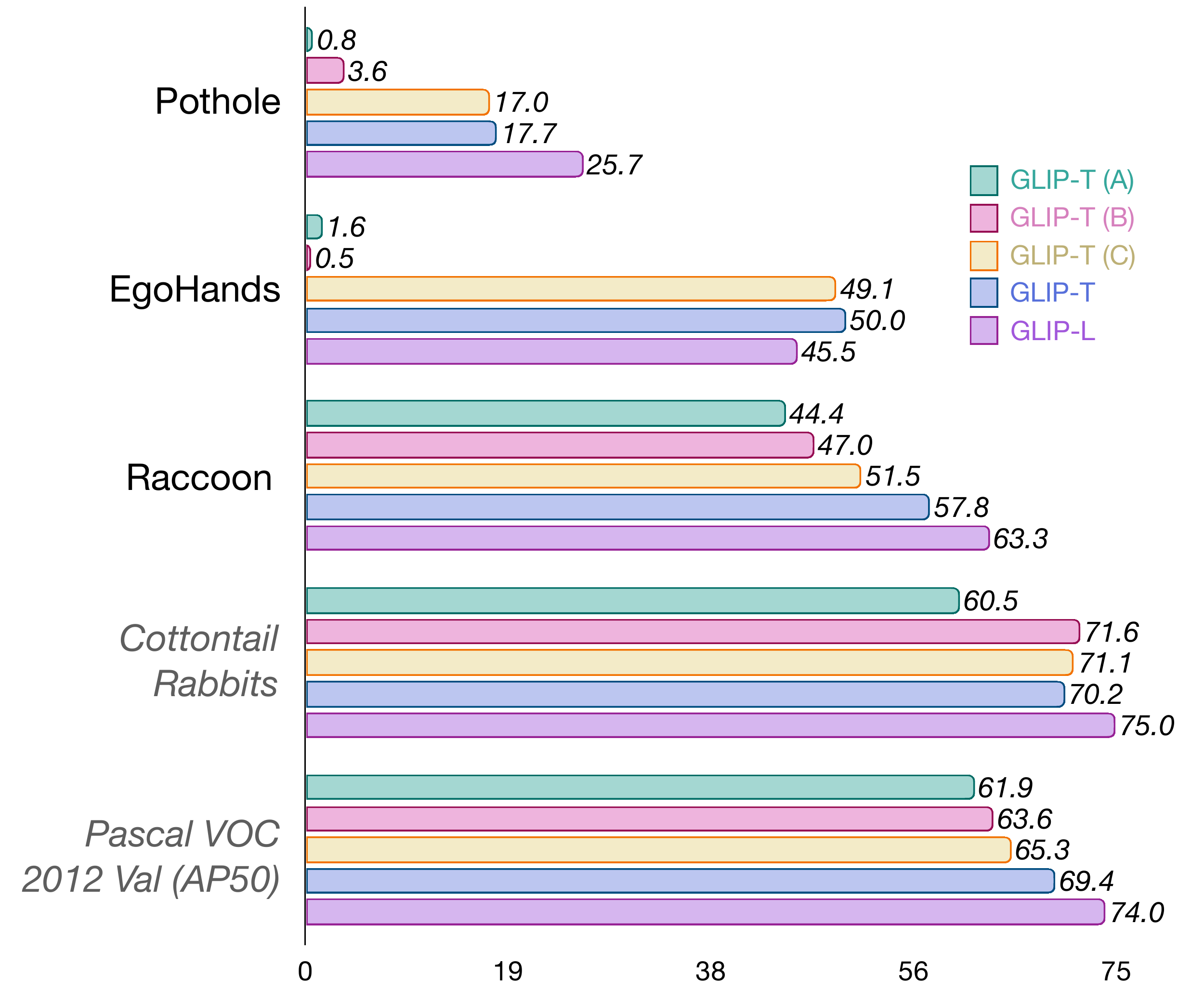}
\caption{Per dataset zero-shot performance. The first 3 datasets contain novel categories not present in the Objects365 vocabulary while the last 2 datasets' categories are covered by Objects365 data. Grounding data bring significant benefit to novel categories.
}
\label{fig:perdataset}
\end{figure}

We further plot the zero-shot performance of GLIP variants on 5 different datasets in Figure \ref{fig:perdataset}. We find that the introduction of grounding data brings significant improvement on certain tasks that test novel concepts, e.g., on Pothole and EgoHands, models without grounding data (A\&B) performs terribly, while models with grounding data (C) outperform them with ease.

\begin{figure}[t]
\centering
\includegraphics[width=\linewidth]{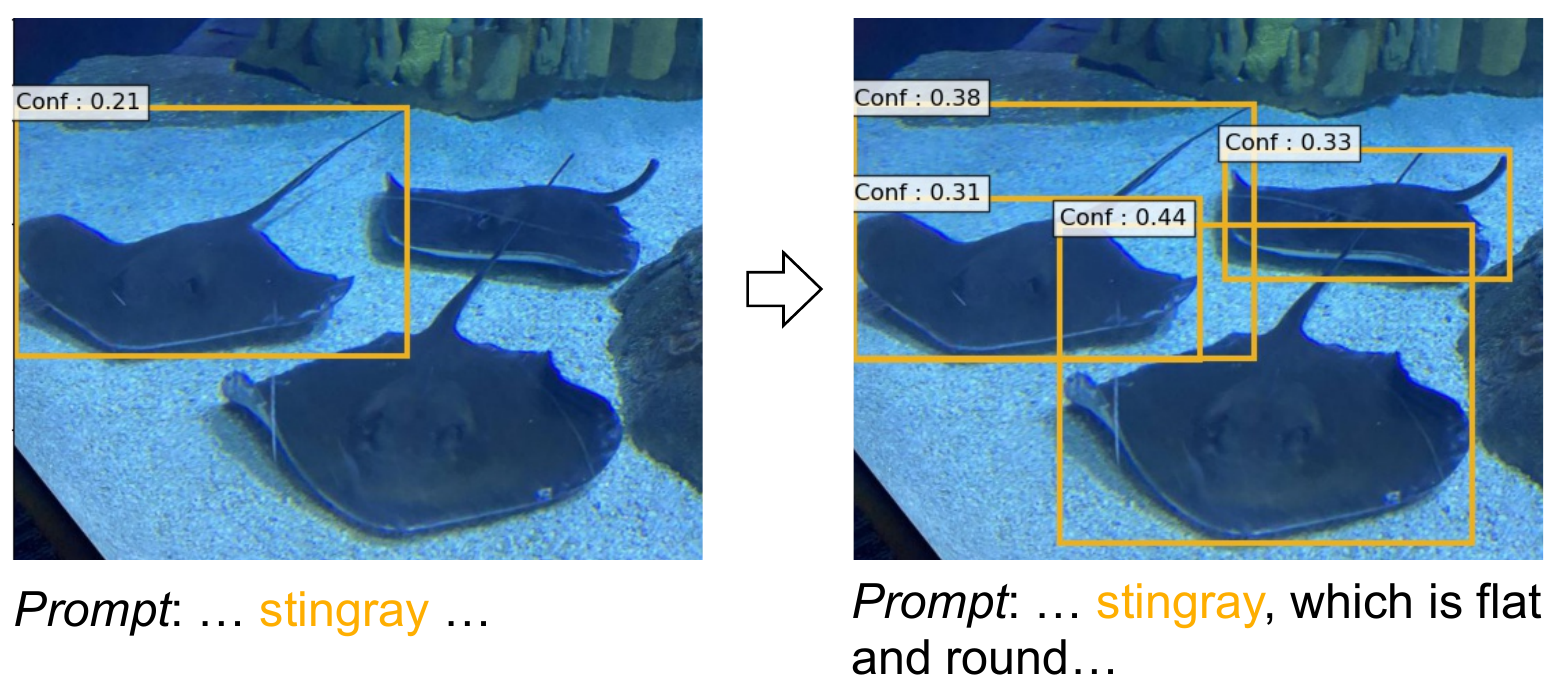}
\caption{A manual prompt tuning example from the Aquarium dataset in ODinW. Given an expressive prompt (``flat and round''), zero-shot GLIP can detect the novel entity ``stingray'' better.}\label{fig:odinw_example} % For simplicity, we show only the predictions for the class ``stingray''.
\vspace{-2mm}
\end{figure}

\subsection{One Model for All Tasks}
\label{sec:onemodel}
As neural models become larger, how to reduce deployment cost has drawn an growing research interest. Recent work on language models \cite{shin2020autoprompt}, image classification \cite{zhou2021learning}, and object detection \cite{wang2020frustratingly} has explored adapting a pre-trained model to a new domain but only changing the least amount of parameters. Such a setting is often denoted as linear probing \cite{kim2018bilinear}, prompt tuning \cite{zhou2021learning}, or efficient task adapters \cite{gao2021clip}. The goal is to have a single model serving various tasks, and each task adds only a few task-specific parameters or no parameters to the pre-trained model. This reduces training and storage cost. In this section, we evaluate models against the metric of deployment efficiency. 

\noindent\textbf{Manual prompt tuning.} 
As \our performs language-aware localization, i.e., the output of GLIP is heavily conditioned on the language input, we propose an efficient way for \our to do task transfer: for any novel categories, the user can use expressive descriptions in the text prompt, adding attributes or language context, to inject domain knowledge and help \our transfer. For example, on the left hand side of Figure \ref{fig:odinw_example}, the model fails to localize all occurrences of the novel entity ``stingray''. However, by adding the attributes to the prompt, i.e., ``flat and round'', the model successfully localizes all occurrences of stringrays. With this simple prompt change, we improve the AP50 on stingray from 4.6 to 9.7. This resembles the prompt design technique in GPT-3 \cite{brown2020language} and is practically appealing, as it requires no annotated data or model re-training. Please refer to the appendix for more details.

\begin{figure}[t]
\centering
\includegraphics[width=0.85\linewidth]{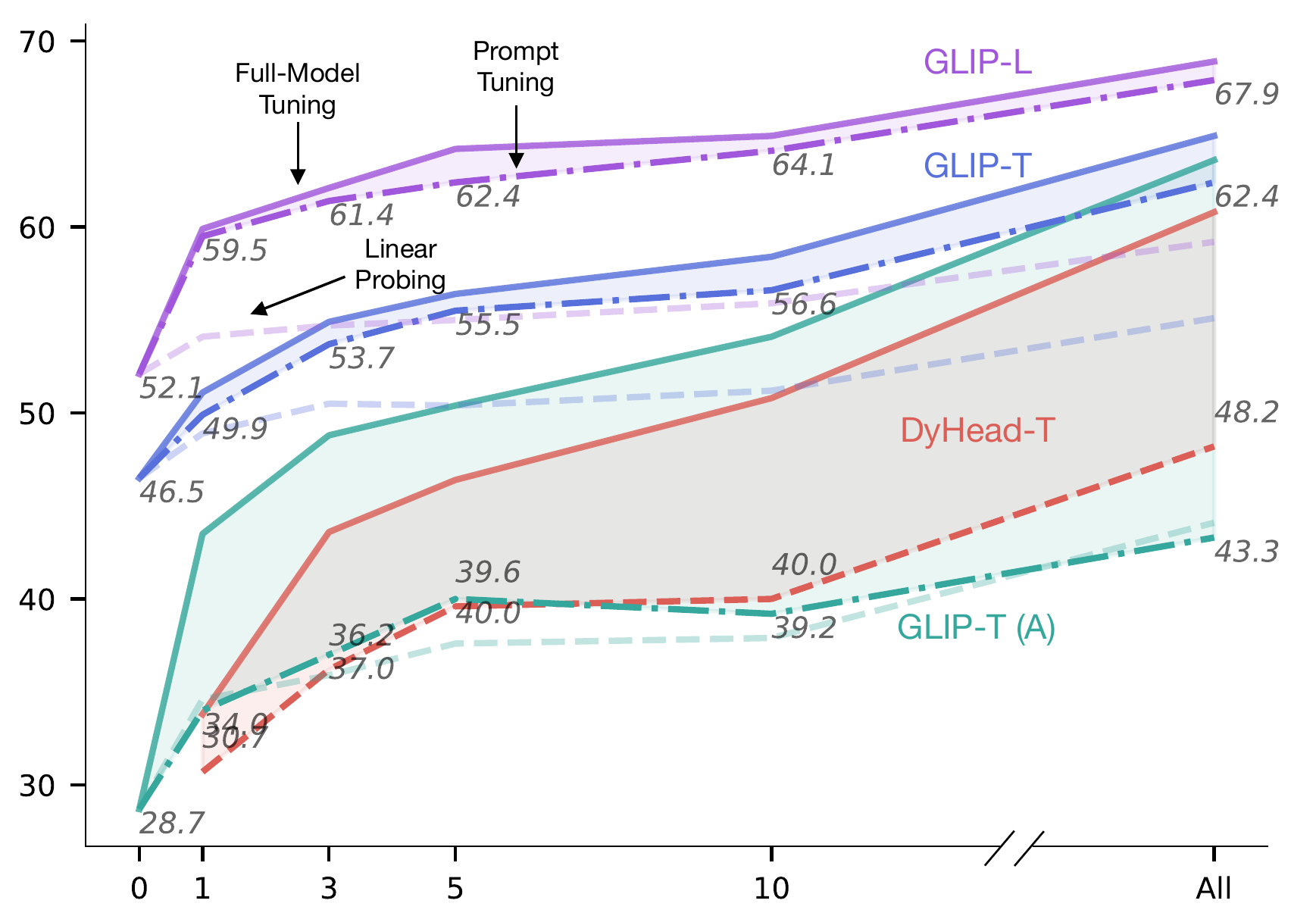}
\caption{Effectiveness of prompt tuning. Solid lines are full-model tuning performance; dashed lines are prompt/linear probing performance.
By only tuning the prompt embeddings, \ourtiny and \ourlarge can achieve performance close to full-model tuning, allowing for efficient deployment.}
\label{fig:prompt}
\vspace{-2mm}
\end{figure}

\noindent\textbf{Prompt tuning.}
We further consider the setting where we have access to task-specific training data but wish to tune the least amount of parameters for easy deployment.
For classical detection models, Wang \etal \cite{wang2020frustratingly} report the effectiveness of ``linear probing'' (i.e., train only the box regression and classification head). \our can also be ``linear probed'', where we only fine-tune the box head and a projection layer between the region and prompt embeddings. 
Because of the language-aware deep fusion, \our supports a more powerful yet still efficient transfer strategy: prompt tuning  \cite{shin2020autoprompt,lester2021power}. 
For GLIP, as each detection task has only one language prompt (e.g., the prompt for Pothole could be ``Detect pothole.'' for all images), we first get prompt embeddings $P^0$ from the language backbone, then discard the language backbone and only fine-tune $P^0$ as the task-specific input (Section \ref{sec:unif_arch}).

We evaluate the models' performance under three settings (Figure \ref{fig:prompt}): linear probing, prompt tuning (only applicable for \our), and full-model tuning. For DyHead-T, prompt tuning is not applicable as the traditional object detection model cannot accept language input; the gap between linear probing and full-model tuning is large. GLIP-T (A) has no language-aware deep fusion; thus prompt tuning and linear tuning achieve similar performance and lag significantly behind full-model tuning. However, for \ourtiny and \ourlarge, prompt tuning almost matches the full-tuning results, without changing any of the grounding model parameters.
Interestingly, as the model and data size grow larger, the gap between full-model tuning and prompt tuning becomes smaller (\ourlarge v.s. \ourtiny), echoing the findings in NLP literature \cite{liu2021pre}. %In Table \ref{table:intro_one_model_for_all}, we report the prompt tuning performance with full data of GLIP-L (along with the prompt-tuning performance on COCO) and the model can achieve high performance on 14 datasets with one set of parameter weights.

\vspace{-1.5mm}
\section{Conclusion}
\vspace{-1.5mm}
GLIP unifies the object detection and phrase grounding tasks to learn an object-level, language-aware, and semantic-rich visual representation. After pre-training, GLIP showed promising results on zero-shot and fine-tuning settings on well-established benchmarks and 13 downstream tasks. We leave a detailed study of how GLIP scales with text-image data size to future work. %is on-going for further understanding.

\vspace{-1.5mm}
\section*{Acknowledgement}
\vspace{-1.5mm}
We thank anonymous reviewers for their comments and suggestions. We thank Xiyang Dai, Zicheng Liu, Yi-Ling Chen for help with the project.
LL and KC are supported in part by DARPA MCS program under Cooperative Agreement N66001-19-2-4032.

%%%%%%%%% REFERENCES
{\small
\bibliographystyle{ieee_fullname}
\bibliography{egbib}
}

\clearpage
\section*{Appendix}

\appendix

This appendix is organized as follows. 

\begin{itemize} %[noitemsep,topsep=0pt,parsep=0pt,partopsep=0pt]
\item In Section \ref{app:visualization}, we provide more visualizations of our model's grounding predictions on the Conceptual Caption 12M dataset \cite{changpinyo2021conceptual}.

\item In Section \ref{app:equivalen} (referred by Section 3.1), we discuss the equivalence between detection and grounding.

\item In Section \ref{app:main_detail} (referred by Section 4), we introduce the pre-training details of the models we use in Section 4.

\item In Section \ref{app:main_detail_eval} (referred by Section 4), we introduce the evaluation details of experiments on COCO, LVIS, and Flickr30K.

\item In Section \ref{app:public_private} (referred by Section 4), we discuss the difference between the public image-text data (Google Conceptual Captions,SBU) and the image-text data we collected.

\item In Section \ref{app:deep_fusion}, we provide a detailed analysis on the computational cost and performance effect of the language-aware deep fusion. 

\item In Section \ref{app:odinw_dataset} (referred by Section 5), we introduce the 13 datasets in Object Detection in the Wild (ODinW).

\item In Section \ref{app:manual_prompt} (referred by Section 5), we detail the manual prompt design.

\item In Section \ref{app:data_efficiency} (referred by Section 5.1), we give the details for the data efficiency experiments.

\item In Section \ref{app:one_model} (referred by Section 5.3), we give the details for the linear probing and prompt tuning experiments.

\item In Section \ref{app:all_results}, we present per-dataset results for all experiments in Section 5.

\end{itemize}
\begin{figure}[h]
\centering
\includegraphics[width=0.95\textwidth]{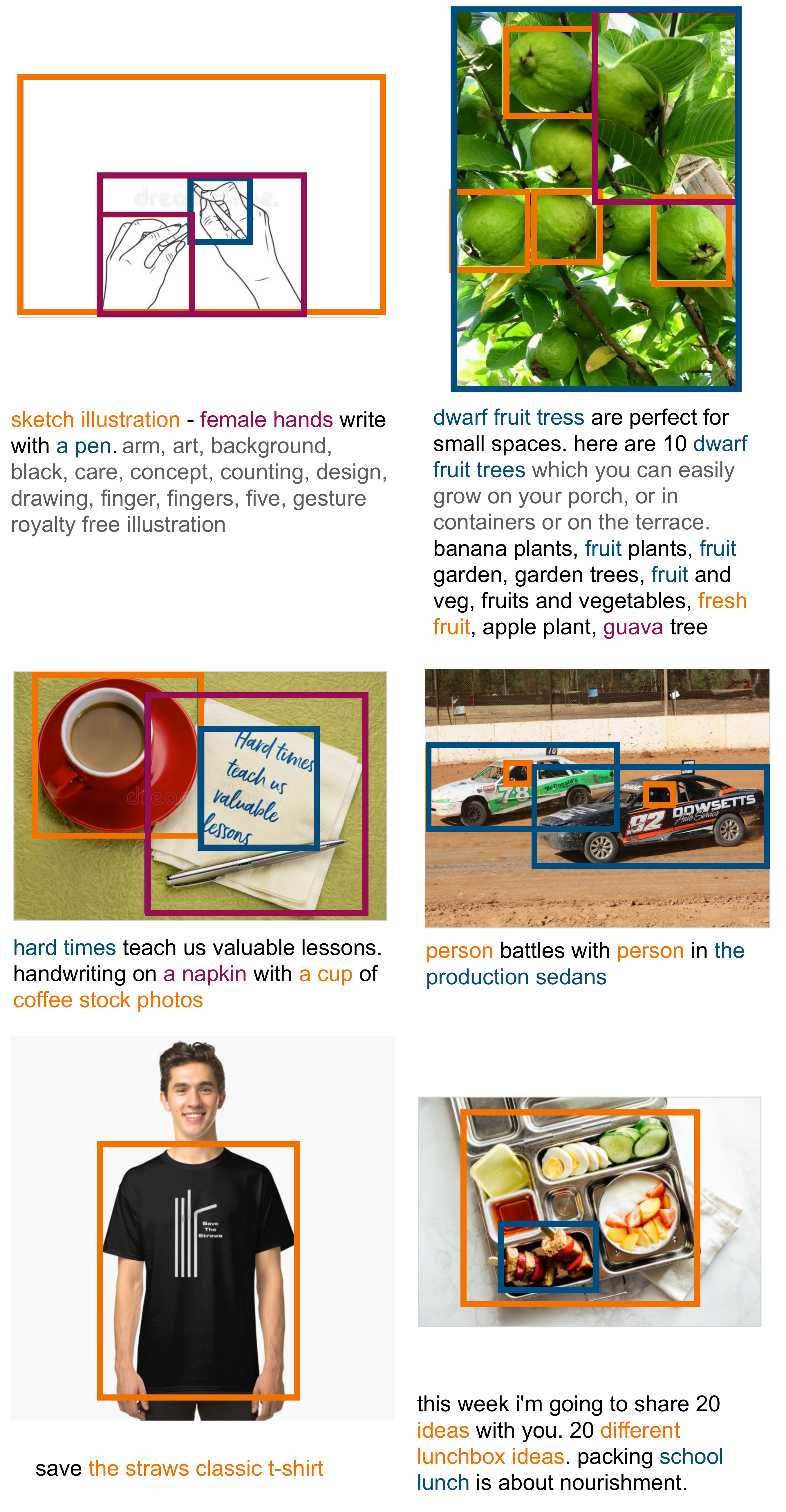}
\caption{Predictions from the teacher model on 6 examples from Conceptual Captions 12M. %The model is trained to predict the root of the phrase as the positive token span. 
Phrases and corresponding boxes are matched with the same colors.
}
\label{fig:moreexamples}
\end{figure}

\section{Visualization}
\label{app:visualization}
We provide more visualizations of the predictions from our teacher model.
Even given noise image-text pairs, our model is still capable of grounding semantic-rich phrases accurately.

\clearpage

\section{Equivalence Discussion between Detection and Grounding}
\label{app:equivalen}

In Section 3.1 of the main paper, we discussed the equivalence between detection and grounding. We corroborate the discussion with empirical experiments.

%\vspace{-2cm}
\paragraph{When all object categories fit into a single prompt.} We first confirm that when all categories fit into one prompt, our grounding formulation is equivalent to classical object detection. We conduct the experiments on COCO \cite{lin2014microsoft}. We first choose the SoTA detection model Dynamic Head (DyHead) \cite{dai2021dynamic} based on the Swin-Tiny Transformer backbone \cite{liu2021swin} as the base object detection mode. We then transform this model into a grounding model as described in Section 3.1: we concatenate the 80 class names with ``. '' into one prompt and replace DyHead's classification loss with our grounding loss. We use BERT (base-uncased) \cite{devlin2018bert} to encode the text prompt. When concatenating the class names, we follow a fixed order.

We train the two models with the exact same hyperparameters as in \cite{dai2021dynamic}: we train with the standard 2x training
configurations \cite{he2017mask}. We train with batch size 32 and learning rate $1\times10^{-4}$ (for the model with grounding reformulation, we use $1\times10^{-5}$ for the BERT text encoder). We decay the learning rate at 67\% and 89\% of the total training steps.

The two models achieve the same performance on COCO 2017val: 49.4 AP. Their results are close to the 49.7 reported in the last row of Table 6 of Dai et al. \cite{dai2021dynamic} (the small difference is presumably due to the implementation difference). Thus, we conclude that when all categories can fit into a single prompt, grounding and detection tasks are equivalent.

%\vspace{-1.9cm}

\paragraph{When not all object categories can fit into a single prompt.} The text encoder for the prompt has a limit on the input sentence length. For example, BERT can only encode sentences containing at most 512 tokens. In our implementation, to reduce computational costs, we limit the input length to 256. 
Thus, for certain datasets with a large vocabulary (e.g., Objects365 \cite{shao2019objects365} has 365 object categories), we cannot fit all category names into one prompt. As a practical solution, we can split the category names into multiple prompts, during both training time and inference time. We find that this incurs minor performance drop. For example, in Table 2 in the main paper, DyHead-T pre-trained on Objects365 achieves 43.6 on COCO zero-shot, while GLIP-T (A) (the grounding reformulated model of DyHead) achieves 42.9 on COCO.

\section{Transfer to Established Benchmarks}

We introduce the implementation details of the models used in Section 4 and discuss the difference between public image-text data and the data crawled by us.

\subsection{Pre-training Details}
\label{app:main_detail}

In Section 4, we introduced GLIP-T (A), GLIP-T (B), GLIP-T (C), GLIP-T, and GLIP-L. We introduce the implementation details in the following.
We pre-train models based on Swin-Tiny models with 32 GPUs and a batch size of 64, and models based on Swin-Large with 64 GPUs and a batch size of 64. We use a base learning rate of $1\times10^{-5}$ for the language backbone and $1\times10^{-4}$ for all other parameters. The learning rate is stepped down by a factor of 0.1 at the 67\% and 89\% of the total training steps. We decay the learning rate when the zero-shot performance on COCO saturates. 
The max input length is 256 tokens for all models.

\begin{table}[t]
\caption{ Comparison between public data and data crawled by us.}
\label{table:public_private}
\begin{center}
\resizebox{\linewidth}{!}{
\begin{tabular}{l@{\hskip9pt} | 
c@{\hskip9pt}c@{\hskip9pt} | c@{\hskip9pt} 
c@{\hskip9pt} c@{\hskip9pt}c@{\hskip9pt} |
c@{\hskip9pt}c@{\hskip9pt}c}
\toprule

\multirow{2}{*}{Pre-Train Data} & 
\multicolumn{2}{c|}{COCO}  &
 \multicolumn{4}{c|}{LVIS minival} &
  \multicolumn{3}{c}{Flickr30K val}
 \\

 & Zero-Shot & Fine-Tune & APr & APc & APf & AP & R@1  & R@5 & R@10  \\

\midrule
O365,GoldG,Cap4M & 46.3 & 54.9 & \textbf{20.8} & \textbf{21.4} & \textbf{31.0} & \textbf{26.0} &  \textbf{85.7} & 95.4 & 96.9  \\

O365,GoldG,CC3M,SBU & \textbf{46.6} & \textbf{55.2} & 20.1 & 21.3 & 31.1 & 25.9 & 85.3 & \textbf{95.7} & \textbf{97.2} \\

\bottomrule
\end{tabular}
}
\end{center}
\end{table}

\paragraph{Prompt design for detection data.}
As noted in Section \ref{app:equivalen}, when we pre-train on datasets such as Objects365, we cannot fit all categories into one prompt. During pre-training, we randomly down-sample the categories and keep only the down-sampled categories in the prompt. We randomly shuffle the categories' order in the prompt. %If a positive category is discarded and not kept in the prompt after down-sampling, we will also drop its corresponding boxes from the box labels. 

The down-sampling is done randomly on the fly for each training example and serves as data augmentation. Specifically, for an example, we denote the positive classes that appear in the image as $C_\text{pos}$ and the rest negative classes as $C_\text{neg}$. We always keep all of $C_\text{pos}$. With a probability of $0.5$, we sample from $C_\text{neg}$ till we have 85 categories in the prompt; with a probability of $0.5$, we uniformly choose an interger $N$ from $[1, 85 - |C_\text{pos}|]$ and put $N$ categories in the prompt.

\paragraph{Augmentation for image-text data with generated boxes.}
When we pre-train the model on image-text data with generated boxes, we find it beneficial to increase the difficulty. We mix a few negative captions (that are from other examples and do not match with the image) with the positive caption (that is matched to the image) to form a longer text input. The model is trained to predict boxes and align them to the correct phrases in the positive caption. The model would need to first identify the positive caption among a few potential captions and then align the box to the correct phrases in the positive caption. This makes the grounding task more challenging and help the model learn a semantic-rich representation during pre-training. This augmentation is also done randomly on the fly. For each training example, with a probability of 0.3, we conduct such augmentation and mix in 19 negative captions; with a probability of 0.3, we mix in a random number (uniformly drawn between 1-19) of negative captions; for the rest of the time, we do not conduct such augmentation.

\subsection{Evaluation Details}
\label{app:main_detail_eval}

For fine-tuning on COCO, we use a base learning rate of $1\times10^{-5}$ for pre-trained models.  

For zero-shot evaluation on LVIS, since LVIS has over 1,000 categories and they cannot be fit into one text prompt, we segment them into multiple chunks, fitting 40 categories into one prompt and query the model multiple times with the different prompts. We find that models tend to overfit on LVIS during the course of pre-training so we monitor the performance on minival for all models and report the results with the best checkpoints.

For zero-shot evaluation on Flickr30K, models may also overfit during the course of pre-training so we monitor the performance on the validation set for all models and report the results with the best checkpoints.

\subsection{Difference Between Public Data and Web-Crawled Data}
\label{app:public_private}

For GLIP-T pre-trained with image-text data, as mentioned in Section 4, we train two versions, one with public data (CC3M,SBU) and another with data we crawled (Cap4M). Here we provide a comparison between the two models in Table \ref{table:public_private}.

The two models differ only slightly, with the Cap4M version better on LVIS while the CC3M+SBU version better on COCO. We conjecture that this is potentially because the public data is more extensively screened and contains more common categories and less rare concepts. Thus it performs slightly better on COCO while lags slightly on LVIS.

\begin{table}[t]
\caption{Computational cost of language-aware deep fusion. For speed, we report FPS, which is the number of images processed per second per GPU (higher is better). For memory consumption, we report the GPU memory used in GB (lower is better). Deep fusion brings less than 1x additional computational cost. }
\label{table:fusion_efficiency}
\begin{center}
\resizebox{\linewidth}{!}{
\begin{tabular}{l@{\hskip9pt} | 
c@{\hskip9pt}|
c@{\hskip9pt}  
c@{\hskip9pt}|c@{\hskip9pt}c@{\hskip9pt}
c@{\hskip9pt}c@{\hskip9pt}c@{\hskip9pt}  c@{\hskip9pt}
c@{\hskip9pt}c@{\hskip9pt}c@{\hskip9pt}c@{\hskip9pt}c@{\hskip9pt}c}
\toprule

\multirow{2}{*}{Model} & \multirow{2}{*}{Fusion} &
\multicolumn{2}{c|}{Inference (P100)}
 & \multicolumn{2}{c}{Train (V100)}\\

  &  & Speed & Memory & Speed & Memory \\
\midrule

\multirow{2}{*}{GLIP-T}  & \xmark  & 4.84 FPS & 1.0 GB & 2.79 FPS & 11.5 GB \\

& \cmark & 2.52 FPS & 2.4 GB &  1.62 FPS & 16.0 GB  \\

\midrule

\multirow{2}{*}{GLIP-L}  & \xmark & 0.54 FPS  & 4.8 GB & 1.27 FPS & 19.7 GB \\

  & \cmark & 0.32 FPS & 7.7 GB & 0.88 FPS & 23.4 GB \\

\bottomrule
\end{tabular}
}
\end{center}
\end{table}

\begin{table*}[t]
\caption{Language-aware fusion benefits most tasks. We reported the full-model tuning performance for ODinW few-shot results. For models trained with only O365, performance on Flickr30K (grey numbers) is significantly worse because the models are not trained to ground natural language captions. }
\label{table:fusion_ablation}
\begin{center}
\resizebox{\linewidth}{!}{
\begin{tabular}{ 
c@{\hskip9pt}c@{\hskip9pt}|
c@{\hskip9pt}  
c@{\hskip9pt}| c@{\hskip9pt}c@{\hskip9pt}
c@{\hskip9pt}c@{\hskip9pt}|c@{\hskip9pt}  c@{\hskip9pt}
c@{\hskip9pt}|c@{\hskip9pt}c@{\hskip9pt}c@{\hskip9pt}c@{\hskip9pt}c@{\hskip9pt}c@{\hskip9pt}c@{\hskip9pt}c}
\toprule

 \multirow{2}{*}{Deep Fusion} &  \multirow{2}{*}{Data} & \multicolumn{2}{c|}{COCO}  &
 \multicolumn{4}{c|}{LVIS minival} &
  \multicolumn{3}{c|}{Flickr30K val} &
  \multicolumn{6}{c}{ODinW}\\

 & & Zero-Shot & Fine-Tune & APr & APc & APf & AP & R@1  & R@5 & R@10 & 0-Shot & 1-Shot & 3-Shot &  5-Shot & 10-Shot & Full-Shot \\ 
\midrule

\xmark & O365  
& 42.9 & 52.9
    & \textbf{14.2} & \textbf{13.9} & \textbf{23.4} & \textbf{18.5} 
    & {\color{gray} 46.4} & {\color{gray} 63.2} & {\color{gray} 66.9}
    & 28.7
& 43.5
& 48.8
& 50.4
& 54.1
& \textbf{63.6}
    \\

\cmark & O365 
    & \textbf{44.9} & \textbf{53.8}
  & 13.5 & 12.8 & 22.2 & 17.8 
  & {\color{gray} 41.4} & {\color{gray} 57.7} & {\color{gray} 61.0}
  &  \textbf{33.2}
& \textbf{48.0}
& \textbf{52.0}
& \textbf{53.2}
& \textbf{54.9}
& 62.7

   \\

\midrule
\xmark & O365,GoldG 
  & 41.6 & 52.9
  & 15.8 & \textbf{23.0} & 30.8 & \textbf{26.1}
  & 82.4 & 94.7 & \textbf{96.6} 
  & 35.5 & 47.2 & 51.9 & 53.8 & 54.3 & \textbf{65.1} 
   \\

\cmark & O365,GoldG 
  & \textbf{46.7} & \textbf{55.1} 
  & \textbf{17.7} & 19.5 & \textbf{31.0} & 24.9
  & \textbf{84.8} & \textbf{94.9} & 96.3
  & \textbf{44.4}
& \textbf{49.6}
& \textbf{53.8}
& \textbf{54.8} 
& \textbf{57.2} 
& 63.9
 \\

\bottomrule
\end{tabular}
}
\end{center}
\end{table*}

\section{Computation Cost and Performance Analysis of Deep Fusion}
\label{app:deep_fusion}

In this section, we provide a more detailed ablation on the computational cost and performance effect of the language-aware deep fusion proposed in Section 3. 

\subsection{Computational Cost}
We test the additional computational cost of the language-aware deep fusion for both GLIP-T and GLIP-L. For inference, we test on a P100 GPU with batch size 1. Note that for inference with GLIP without deep fusion, we could cache the language embeddings of the prompts; thus the inference time of GLIP without deep fusion is equivalent to that of DyHead \cite{dai2021dynamic}. 

For training, we test on a standard DGX-2 machine with 16 V100 GPUs (we test under the multi-GPU setting as it mimics the actual training environment): for GLIP-T models, we use 2 images per batch and for GLIP-L models, we use 1 images per batch. As the fusion module invovles multi-head attention over a large number of input elements, we turn on gradient checkpointing\footnote{\url{https://pytorch.org/docs/stable/checkpoint.html}} for the deep fusion module, which increases training time but reduces GPU memory consumption. 

Table \ref{table:fusion_efficiency} shows that the language-aware deep fusion brings less than 1x additional computational cost overall. %With gradient checkpointing, the training memory consumption increases less than $50\%$ while the  

\subsection{Performance}

We provide an analysis on the effect of language-aware deep fusion when different kinds of pre-training data are used. We pre-train four variants of GLIP-T and show the results In Table \ref{table:fusion_ablation}. Deep fusion is beneficial for testing on 1) common categories (i.e., COCO); 2) grounding tasks (i.e., Flickr30K), and 3) low-resource transfer to real-world downstream tasks (i.e., ODinW). 

However, on LVIS, the effect of deep fusion seems unclear: when only detection data are used, deep fusion seems to degrades performance (row 1 v.s. row 2); when grounding data are present, deep fusion degrades common category performance but improves rare category performance. Our assumption is that when GLIP is only trained with detection data (e.g., O365), the language model could ``overfit'' to the categories in O365 and does not generalize to novel categories well (i.e., outputs out-of-distribution text representation). The deep fusion could ``amplify'' such overfit as the visual representation is conditioned on the language model. Thus, when tested on prompts containing novel categories (e.g., LVIS), deep fusion could degrade performance. When grounding data are used, such overfit could be mitigated. %Finally, early fusion is crucial in learning prompt-conditioned image features (see Figure 5 and Sec 5.2).

\section{Object Detection in the Wild}
In this section, we provide the details and additional results for the experiments in Section 5.

\subsection{Dataset Details}
\label{app:odinw_dataset}
We use 13 datasets from Roboflow\footnote{\url{https://public.roboflow.com/object-detection}}. Roboflow hosts over 30 datasets and we exclude datasets that are too challenging (e.g., detecting different kinds of chess pieces) or impossible to solve without specific domain knowledge (e.g., understanding sign language).

We provide the details of the 13 datasets we use in Table \ref{table:odinw_dataset}.  We include the PASCAL V0C 2012 dataset as a reference dataset, as public baselines have been established on this dataset.
 For PascalVOC, we follow the convention and report on validation set. For Pistols, there are no official validation or test sets so we split the dataset ourselves. 

\begin{table*}[t]
\caption{13 ODinW dataset statistics. We summarize the objects of interest for each dataset and report the image number of each split. }
\label{table:odinw_dataset}
\begin{center}
\resizebox{\linewidth}{!}{
\begin{tabular}{l@{\hskip9pt} | 
c@{\hskip9pt}|c@{\hskip9pt}|
c@{\hskip9pt}  
c@{\hskip9pt} c@{\hskip9pt}c@{\hskip9pt}
c@{\hskip9pt}c@{\hskip9pt}c@{\hskip9pt}  c@{\hskip9pt}
c@{\hskip9pt}c@{\hskip9pt}c@{\hskip9pt}c@{\hskip9pt}c@{\hskip9pt}c}
\toprule

Dataset & Objects of Interest & Train/Val/Test & URL \\
\midrule
PascalVOC & Common objects (PascalVOC 2012) & 13690/3422/- & \url{https://public.roboflow.com/object-detection/pascal-voc-2012} \\
AerialDrone & Boats, cars, etc. from drone images & 52/15/7 &  \tiny{\url{https://public.roboflow.com/object-detection/aerial-maritime}}\\
Aquarium & Penguins, starfish, etc. in an aquarium & 448/127/63 & \tiny{\url{https://public.roboflow.com/object-detection/aquarium}} \\
Rabbits & Cottontail rabbits & 1980/19/10 & \tiny{\url{https://public.roboflow.com/object-detection/cottontail-rabbits-video-dataset}} \\
EgoHands & Hands in ego-centric images & 3840/480/480 & \tiny{\url{https://public.roboflow.com/object-detection/hands}} \\
Mushrooms & Two kinds of mushrooms & 41/5/5 &  \url{https://public.roboflow.com/object-detection/na-mushrooms}\\
Packages & Delivery packages & 19/4/3 & \url{https://public.roboflow.com/object-detection/packages-dataset} \\

Raccoon & Raccoon & 150/29/17 & \url{https://public.roboflow.com/object-detection/raccoon} \\

Shellfish & Shrimp, lobster, and crab & 406/116/58 & \url{https://public.roboflow.com/object-detection/shellfish-openimages} \\

Vehicles & Car, bus, motorcycle, truck, and ambulance & 878/250/126 & \url{https://public.roboflow.com/object-detection/vehicles-openimages} \\

Pistols & Pistol & 2377/297/297 & \url{https://public.roboflow.com/object-detection/pistols/1}\\

Pothole & Potholes on the road & 465/133/67 &  \url{https://public.roboflow.com/object-detection/pothole} \\

Thermal & Dogs and people in thermal images & 142/41/20 & \url{https://public.roboflow.com/object-detection/thermal-dogs-and-people} \\

\bottomrule
\end{tabular}
}
\end{center}
\end{table*}

\subsection{Manual Prompt Tuning}
\label{app:manual_prompt}
As discussed in Section 5, we find it beneficial to manually design some prompts to provide language guidance. We provide the prompts we use in Table \ref{table:prompt_design}. We design the prompts for 6 datasets. Since some prompts are sentences, we only apply these prompts for models trained with grounding data (GLIP-T (C), GLIP-T, and GLIP-L). For GLIP-T (A) and GLIP-T (B), we find it beneficial to use prompts for the Rabbits and Mushrooms datasets, as the prompts there are just single word or short phrases. Overall, using prompts improves AP without any model re-training (e.g., the AP improves from 22.1 to 50.0 for EgoHands).

\begin{table}[t]
\caption{Manually designed prompts for 6 datasets. Words in \textit{italic} are the objects of interest. The prompts either provide attributes, specify the category name in more common words, or provide language contexts. They can improve AP (CLIP-T) without any annotation or model re-training. Specifically for Pothole, although the changed prompt does not improve the AP of CLIP-T, we find it effective for CLIP-T (C) so we still apply the prompt.}
\label{table:prompt_design}
\begin{center}
\resizebox{\linewidth}{!}{
\begin{tabular}{l@{\hskip9pt} | 
c@{\hskip9pt}c@{\hskip9pt}|
c@{\hskip9pt}  
c@{\hskip9pt} c@{\hskip9pt}c@{\hskip9pt}
c@{\hskip9pt}c@{\hskip9pt}c@{\hskip9pt}  c@{\hskip9pt}
c@{\hskip9pt}c@{\hskip9pt}c@{\hskip9pt}c@{\hskip9pt}c@{\hskip9pt}c}
\toprule

Dataset & Original Prompt & AP &  Manually Designed Prompts & AP \\
\midrule

\multirow{3}{*}{Aquarium}  & 

\textit{penguin} & \multirow{3}{*}{17.7}   & \textit{penguin}, which is black and white & \multirow{3}{*}{18.4} \\
& \textit{puffin} & & \textit{puffin} with orange beaks &  \\ 
& \textit{stingray} & & \textit{stingray} which is flat and round & \\

\midrule
Rabbits & \textit{Cottontail-Rabbits} & 68.0 & \textit{rabbit} & 70.2 \\\midrule
EgoHands & \textit{hand} & 22.1 & \textit{hand} of a person  & 50.0 \\\midrule
Mushrooms & \textit{Cow. Chanterelle} & 13.6 & \textit{flat mushroom}. \textit{yellow mushroom} & 73.8 \\\midrule
Packages & \textit{package} & 50.0 & there is a \textit{package} on the porch & 72.3 \\\midrule

Pothole & \textit{pothole} & 17.8 & there are some \textit{holes} on the road & 17.7 \\

\bottomrule
\end{tabular}
}
\end{center}
\end{table}

\begin{table}[t]
\caption{ Zero-shot and full fine-tuning performance. GLIP models exhibit superior data efficiency.}
\label{table:full_tuning}
\begin{center}
\resizebox{\linewidth}{!}{
\begin{tabular}{l@{\hskip9pt} | 
c@{\hskip9pt}|c@{\hskip9pt}c@{\hskip9pt} 
c@{\hskip9pt} c@{\hskip9pt}c@{\hskip9pt}
c@{\hskip9pt}c@{\hskip9pt}c@{\hskip9pt}  c@{\hskip9pt}
c@{\hskip9pt}c@{\hskip9pt}c@{\hskip9pt}c@{\hskip9pt}c@{\hskip9pt}c}
\toprule

\multirow{2}{*}{Model} & 
\multirow{2}{*}{Zero Shot} &
 \multicolumn{5}{c}{Full Tuning}\\

&  & 1 & 3 & 5 & 10 & All \\

\midrule

DyHead-T \scriptsize{COCO}
    & -

& 31.9\std{4.1}
& 44.2\std{0.4}
& 44.7\std{2.1}
& 50.1\std{2.0}
& 63.2

\\ % from collection setv2_nov12

DyHead-T  \scriptsize{O365}  
    & -

& 33.8\std{4.3}
& 43.6\std{1.2}
& 46.4\std{1.4}
& 50.8\std{1.6}
& 60.8
\\

\midrule

GLIP-T (A)  & 28.7
& 43.5\std{1.5}
& 48.8\std{0.4}
& 50.4\std{0.7}
& 54.1\std{0.5}
& 63.6
\\

GLIP-T (B)  &  33.2 
& 48.0\std{0.8}
& 52.0\std{0.4}
& 53.2\std{0.9}
& 54.9\std{0.7}
& 62.7

  \\
  
GLIP-T (C)  &  44.4

& 49.6\std{0.3}
& 53.8\std{0.2}
& 54.8\std{1.0}
& 57.2\std{1.1}
& 63.9

  \\

GLIP-T  & 46.5

& 51.1\std{0.1}
& 54.9\std{0.3}
& 56.4\std{0.5}
& 58.4\std{0.2}
& 64.9
\\

\midrule

GLIP-L & 52.1

& 59.9\std{1.7}
& 62.1\std{0.8}
& 64.2\std{0.4}
& 64.9\std{0.9}
& 68.9

\\

\bottomrule
\end{tabular}
}
\end{center}
\end{table}

\subsection{Data Efficiency}
\label{app:data_efficiency}

We provide details for the experiments in Section 5.1. We train with batch size 4, learning rate $1\times10^{-4}$ (for the model with grounding reformulation, we use $1\times10^{-5}$ for the BERT text encoder), and weight decay of 0.05. We do not find that increasing batch size improves performance significantly. For computational reasons, we use a batch size of 4. Following convention, we freeze the bottom 2 layers of the backbone during fine-tuning. We monitor the performance on validation and decay the learning rate by 0.1 when the validation performance plateaus. In $X$-shot settings, we randomly sample the dataset such that there are at least $X$ examples per category \cite{kang2019few}. We change the random seeds (and thus change the sampled data) and conduct 3 independent runs for each $X$-shot experiment. We provide two DyHead-T variants as baselines, one trained on COCO and one trained on Objects365. We report the full zero-shot results in Table \ref{table:zero_shot_full} and few-shot results in Table \ref{table:full_tuning}.

\begin{table}[t]
\caption{Linear probing performance.}
\label{table:linear}
\begin{center}
\resizebox{\linewidth}{!}{
\begin{tabular}{l@{\hskip9pt} | 
c@{\hskip9pt}c@{\hskip9pt}c@{\hskip9pt} 
c@{\hskip9pt}c@{\hskip9pt}c@{\hskip9pt}
c@{\hskip9pt}c@{\hskip9pt}c@{\hskip9pt}  c@{\hskip9pt}
c@{\hskip9pt}c@{\hskip9pt}c@{\hskip9pt}c@{\hskip9pt}c@{\hskip9pt}c}
\toprule

\multirow{2}{*}{Model} & 
\multicolumn{5}{c}{Linear Probing}\\

 & 1 & 3 & 5 & 10 & All \\

\midrule

DyHead-T \scriptsize{COCO}
& 22.7\std{1.1}
& 32.7\std{1.4}
& 30.5\std{2.9}
& 34.1\std{1.4}
& 43.1

\\

DyHead-T \scriptsize{COCO-Cosine}  
& 21.8\std{4.4}
& 30.6\std{2.2}
& 33.3\std{1.2}
& 35.5\std{1.2}
& 43.5

\\

DyHead-T   \scriptsize{O365}  
& 30.7\std{3.3} 
& 36.2\std{3.3}
& 39.6\std{0.4}
& 40.0\std{2.7}
& 48.2

\\
DyHead-T  \scriptsize{O365-Cosine}  
& 25.2\std{2.6}
& 37.6\std{0.5}
& 38.9\std{0.7}
& 41.5\std{0.5}
& 49.4

\\

\midrule

 GLIP-T (A)
& 34.6\std{0.7}
& 35.9\std{0.2}
& 37.6\std{0.1}
& 37.9\std{0.2}
& 44.1 
\\

  GLIP-T (B) 
& 40.9\std{0.3}
& 42.8\std{0.4}
& 44.0\std{0.2}
& 44.4\std{0.3}
& 51.8

  \\
  
  GLIP-T (C)
& 43.9\std{0.1}
& 45.4\std{0.1}
& 45.9\std{0.2}
& 46.7\std{0.3}
& 52.7

  \\

 GLIP-T  
& 48.9\std{0.2}
& 50.5\std{0.1}
& 50.4\std{0.3}
& 51.2\std{0.2}
& 55.1

  \\
\midrule

  GLIP-L
& 54.1\std{0.3}
& 54.7\std{0.2}
& 55.0\std{0.0}
& 55.9\std{0.4}
& 59.2

\\

\bottomrule
\end{tabular}
}
\end{center}
\end{table}

\begin{table}[t]
\caption{Prompt tuning performance.}
\label{table:prompt_tuning}
\begin{center}
\resizebox{\linewidth}{!}{
\begin{tabular}{l@{\hskip9pt} | 
c@{\hskip9pt}c@{\hskip9pt}c@{\hskip9pt} 
c@{\hskip9pt} c@{\hskip9pt}c@{\hskip9pt}
c@{\hskip9pt}c@{\hskip9pt}c@{\hskip9pt}  c@{\hskip9pt}
c@{\hskip9pt}c@{\hskip9pt}c@{\hskip9pt}c@{\hskip9pt}c@{\hskip9pt}c}
\toprule

\multirow{2}{*}{Model} & 
 \multicolumn{5}{c}{Prompt Probing}\\

 & 1 & 3 & 5 & 10 & All  \\

\midrule

GLIP-T (A)
& 34.0\std{0.1}
& 37.0\std{0.6}
& 40.0\std{0.4}
& 39.2\std{1.0}
& 43.3
\\

GLIP-T (B)
& 46.4\std{0.5}
& 49.0\std{0.9}
& 50.6\std{0.5}
& 52.7\std{0.1}
& 58.5
  \\
  
GLIP-T (C)

& 50.6\std{0.5}
& 52.9\std{0.5}
& 53.9\std{0.7}
& 55.8\std{1.1}
& 62.8 

  \\

GLIP-T 

& 49.9\std{0.7}
& 53.7\std{1.6}
& 55.5\std{0.6}
& 56.6\std{0.3}
& 62.4
  \\
% GLIP-T

% & 49.5\std{0.6}
% & 53.5\std{0.4}
% & 55.3\std{0.5}
% & 57.8\std{0.2}
% & 63.1 \\

GLIP-L
& 59.5\std{0.4}
& 61.4\std{0.4}
& 62.4\std{0.6}
& 64.1\std{0.6} 
& 67.9
\\
\bottomrule
\end{tabular}
}
\end{center}
\end{table}

\subsection{One Model for All Tasks}
\label{app:one_model}

In Section 5.2, we conduct experiments with respect to deployment efficiency: tuning the least amount of parameters for the best performance. For all models, we experiment with the linear probing setting; for GLIP models, we also experiment with the prompt tuning setting. For linear probing, we try both the vanilla approach (simply tune the classification and localization head) and the cosine scale approach \cite{wang2020frustratingly}. Below we provide the implementation details.

For the vanilla linear probing, we train with a learning rate of $1\times10^{-4}$, batch size of 4, and weight decay of 0.05. For linear probing with the cosine scale, we use a scale of $20.0$ per suggestions of Wang et al. \cite{wang2020frustratingly}, learning rate of $0.01$, batch size of 4, and weight decay of 0.05. For prompt tuning, we train with a learning rate of $0.05$, batch size of 4, and weight decay of 0.25. We have conducted preliminary searches for the hyper-parameters.

Results are present in Table \ref{table:linear} (linear probing) and Table \ref{table:prompt_tuning} (prompt tuning). Comparing them with full-tuning results (Table \ref{table:full_tuning}), we see prompt tuning performance of GLIP is competitive, showing the deployment efficiency. Contrary to Wang \etal \cite{wang2020frustratingly} who report that linear probing can deliver competitive performance for classical detection models, we find that linear probing does not work well compared to full tuning. We find that the reason could be the transfer datasets (ODinW) in our case contain a lot of novel tasks and domains, while experiments in Wang \etal focus on transferring to common domains (e.g., PascalVOC and COCO). In Table \ref{table:perdataset_all_1}, we report the per-dataset performance. We find that for some common tasks or domains (e.g., PascalVOC and Vehicles), linear probing of DyHead COCO performs competitively with full fine-tuning but the gap is large for some other tasks of a novel domain (e.g., AerialDrone).

\subsection{All Results}
\label{app:all_results}
We report the per-dataset performance under 0,1,3,5,10-shot and full data as well as linear probing, prompt tuning, and full-model tuning in Table \ref{table:zero_shot_full}, Table \ref{table:perdataset_all_1}, and Table \ref{table:perdataset_all_2} (on the next pages).
\clearpage

\begin{table*}[ht]
\caption{Zero-shot performance on 13 ODinW datasets.}
\label{table:zero_shot_full}
\begin{center}
\resizebox{\linewidth}{!}{
\begin{tabular}{l@{\hskip9pt}| 
l@{\hskip9pt}l@{\hskip9pt}l@{\hskip9pt} 
l@{\hskip9pt}l@{\hskip9pt}l@{\hskip9pt}
l@{\hskip9pt}l@{\hskip9pt}l@{\hskip9pt}l@{\hskip9pt}
l@{\hskip9pt}l@{\hskip9pt}l@{\hskip9pt}l@{\hskip9pt}l@{\hskip9pt}l@{\hskip9pt}l}
\toprule

Model  & \small{PascalVOC} &
\small{AerialDrone} & 
\small{Aquarium} &
\small{Rabbits} &
\small{EgoHands} &
\small{Mushrooms} &
\small{Packages} &
\small{Raccoon} &
\small{Shellfish} &
\small{Vehicles} &
\small{Pistols} &
\small{Pothole} &
\small{Thermal} & 
Avg
\\
\midrule
 GLIP-T (A) 
 & 47.7 % 61.9
& 9.8
& 16.8
& 60.5
& 1.6
& 13.7
& 48.5
& 44.4
& 20.4
& 52.4
& 25.3
& 0.8
& 32.3
  & 28.8

 \\

 GLIP-T (B)
 & 50.6 % 63.6
& 4.9
& 19.4
& 71.6
& 0.5
& 21.8
& 29.7
& 47.0
& 21.4
& 56.0
& 47.4
& 3.6
& 57.1
  & 33.2
 \\

 GLIP-T (C)
 & 51.6 % 65.3
& 8.1
& 22.6
& 71.1
& 49.1
& 69.4
& 65.6
& 51.5
& 29.3
& 49.9
& 42.7
& 17.0
& 49.2
  & 44.4\\
 GLIP-T % 69.4
 & 56.2
& 12.5
& 18.4
& 70.2
& 50.0
& 73.8
& 72.3
& 57.8
& 26.3
& 56.0
& 49.6
& 17.7
& 44.1
  & 46.5
  
\\

 GLIP-L  % 74.0
 & 61.7
& 7.1
& 26.9
& 75.0
& 45.5
& 49.0
& 62.8
& 63.3
& 68.9
& 57.3
& 68.6
& 25.7
& 66.0
& 52.1
\\

\bottomrule
\end{tabular}
}
\end{center}
\end{table*}

\begin{table*}[h]
\caption{Per-dataset performance of DyHead, GLIP-T, and GLIP-L. For PascalVOC, we report the mAP (IoU=0.50:0.95) using the COCO evaluation script, to be consistent with other 12 datasets. ``Linear'' denotes linear probing. ``Prompt'' denotes prompt tuning. ``Full'' denotes full-model tuning.}
\label{table:perdataset_all_1}
\begin{center}
\resizebox{\linewidth}{!}{
\begin{tabular}{l@{\hskip9pt} 
c@{\hskip9pt}c@{\hskip9pt}|l@{\hskip9pt} 
l@{\hskip9pt}l@{\hskip9pt}l@{\hskip9pt}
l@{\hskip9pt}l@{\hskip9pt}l@{\hskip9pt}l@{\hskip9pt}
l@{\hskip9pt}l@{\hskip9pt}l@{\hskip9pt}l@{\hskip9pt}l@{\hskip9pt}l}
\toprule

Model & Shot & Tune & \small{PascalVOC} &
\small{AerialDrone} & 
\small{Aquarium} &
\small{Rabbits} &
\small{EgoHands} &
\small{Mushrooms} &
\small{Packages} &
\small{Raccoon} &
\small{Shellfish} &
\small{Vehicles} &
\small{Pistols} &
\small{Pothole} &
\small{Thermal} & 
Avg
\\\midrule
DyHead \scriptsize{COCO} & 1 & Linear 
& 48.2\std2.4
& 2.7\std2.0
& 8.5\std1.5
& 57.8\std3.2
& 9.7\std3.4
& 30.2\std18.3
& 13.2\std9.4
& 30.2\std4.0
& 9.9\std4.0
& 42.5\std4.1
& 5.7\std7.1
& 2.6\std2.0
& 34.2\std19.7
& 22.7\std{0.9}

\\ 
DyHead \scriptsize{COCO} & 3 & Linear 
& 55.6\std0.6
& 2.7\std3.0
& 12.3\std0.5
& 57.4\std3.1
& 15.4\std2.1
& 57.1\std1.6
& 30.6\std16.9
& 55.4\std1.6
& 14.8\std1.4
& 51.0\std3.9
& 22.8\std3.1
& 8.7\std1.0
& 41.5\std11.1
& 32.7\std{1.1}
\\
DyHead \scriptsize{COCO} & 5 & Linear 
& 56.4\std0.2
& 2.7\std2.4
& 14.1\std0.9
& 54.7\std4.9
& 8.8\std6.6
& 47.1\std12.6
& 24.6\std22.9
& 51.6\std2.9
& 17.0\std0.6
& 46.6\std3.0
& 20.3\std13.9
& 7.8\std2.1
& 44.3\std4.2
& 30.5\std{2.4}
\\
DyHead \scriptsize{COCO} & 10 & Linear 

& 57.4\std0.3
& 7.4\std0.7
& 16.0\std2.2
& 59.8\std0.8
& 18.6\std0.3
& 55.0\std0.8
& 30.8\std17.1
& 53.0\std4.0
& 16.7\std0.7
& 50.7\std0.9
& 27.8\std1.9
& 3.1\std4.3
& 47.5\std3.1
& 34.1\std{1.2}
\\
DyHead \scriptsize{COCO} & All & Linear 

& 61.3
& 10.3
& 21.6
& 61.4
& 39.0
& 55.4
& 54.4
& 57.3
& 23.1
& 60.7
& 47.9
& 14.9
& 53.5
& 43.1
\\
\midrule
DyHead \scriptsize{COCO} & 1 & Full 
& 31.7\std3.1
& 14.3\std2.4
& 13.1\std2.0
& 63.6\std1.4
& 40.9\std7.0
& 67.0\std3.6
& 34.6\std12.1
& 45.9\std3.8
& 10.8\std5.0
& 34.0\std3.3
& 12.0\std10.4
& 6.1\std1.3
& 40.9\std7.4
& 31.9\std{3.3}

\\
DyHead \scriptsize{COCO} & 3 & Full
& 44.1\std0.7
& 19.2\std3.0
& 22.6\std1.3
& 64.8\std1.7
& 54.4\std2.5
& 78.9\std1.3
& 61.6\std10.3
& 50.0\std2.1
& 20.8\std3.5
& 44.9\std1.9
& 34.4\std11.1
& 20.6\std2.4
& 57.9\std2.3
& 44.2\std{0.3} 
\\
DyHead \scriptsize{COCO} & 5 & Full

& 44.9\std1.5
& 22.2\std3.0
& 31.7\std1.0
& 65.2\std1.5
& 55.6\std3.7
& 78.7\std3.9
& 50.1\std13.7
& 48.7\std4.8
& 22.8\std3.3
& 52.0\std1.2
& 39.8\std6.7
& 20.9\std1.5
& 48.0\std2.8
& 44.7\std{1.7}

\\
DyHead \scriptsize{COCO} & 10 & Full

& 48.4\std1.2
& 27.5\std1.4
& 39.3\std2.7
& 62.1\std5.9
& 61.6\std1.4
& 81.7\std3.4
& 58.8\std9.0
& 52.9\std3.2
& 30.1\std3.2
& 54.1\std3.3
& 44.8\std4.9
& 26.7\std2.4
& 63.4\std2.8
  & 50.1\std{1.6}
  \\
DyHead \scriptsize{COCO} & All & Full
& 60.1
& 27.6
& 53.1
& 76.5
& 79.4
& 86.1
& 69.3
& 55.2
& 44.0
& 61.5
& 70.6
& 56.6
& 81.0
  & 63.2

\\
\midrule
\midrule
DyHead \scriptsize{O365} & 1 & Linear
& 45.2\std{3.0}
& 10.8\std{3.6}
& 13.8\std{0.7}
& 61.4\std{0.7}
& 8.9\std{6.3}
& 52.6\std{8.7}
& 58.7\std{3.7}
& 44.0\std{10.4}
& 14.9\std{2.9}
& 40.0\std{0.4}
& 6.9\std{5.0}
& 1.7\std{1.2}
& 39.8\std{7.2}
  & 30.7\std{2.7}
  
 \\
 
 DyHead \scriptsize{O365} & 3 & Linear
& 54.6\std{0.4}
& 12.4\std{3.0}
& 22.3\std{1.5}
& 64.0\std{2.4}
& 10.5\std{6.8}
& 53.6\std{10.6}
& 49.1\std{16.3}
& 60.5\std{1.6}
& 20.6\std{2.2}
& 51.3\std{2.3}
& 25.5\std{0.9}
& 8.2\std{1.1}
& 38.9\std{12.6}
  & 36.3\std{2.7}
\\

 DyHead \scriptsize{O365} & 5 & Linear
& 56.1\std{0.4}
& 13.6\std{1.8}
& 24.8\std{1.1}
& 63.1\std{5.5}
& 15.3\std{1.6}
& 55.2\std{10.3}
& 70.2\std{2.8}
& 60.1\std{2.4}
& 23.0\std{1.4}
& 53.5\std{0.9}
& 26.1\std{2.1}
& 6.8\std{2.3}
& 46.9\std{3.5}
  & 39.6\std{0.4}
\\

 DyHead \scriptsize{O365} & 10 & Linear
& 57.5\std{0.3}
& 8.2\std{3.0}
& 28.2\std{0.8}
& 65.4\std{3.2}
& 17.5\std{0.6}
& 68.0\std{0.8}
& 49.8\std{17.3}
& 60.3\std{2.1}
& 22.9\std{1.0}
& 56.4\std{0.8}
& 28.0\std{2.2}
& 7.6\std{0.9}
& 50.3\std{0.5}
  & 40.0\std{2.2}
\\

 DyHead \scriptsize{O365} & All & Linear
& 63.0 
& 18.9 
& 33.7 
& 69.2 
& 36.3 
& 70.9 
& 52.4 
& 66.7 
& 26.6 
& 60.6 
& 48.2 
& 16.1 
& 64.6 
  & 48.2
\\
\midrule
 DyHead \scriptsize{O365} & 1 & Full
& 25.8\std{3.0}
& 16.5\std{1.8}
& 15.9\std{2.7}
& 55.7\std{6.0}
& 44.0\std{3.6}
& 66.9\std{3.9}
& 54.2\std{5.7}
& 50.7\std{7.7}
& 14.1\std{3.6}
& 33.0\std{11.0}
& 11.0\std{6.5}
& 8.2\std{4.1}
& 43.2\std{10.0}
  & 33.8\std{3.5}
\\
 DyHead \scriptsize{O365} & 3 & Full
& 40.4\std{1.0}
& 20.5\std{4.0}
& 26.5\std{1.3}
& 57.9\std{2.0}
& 53.9\std{2.5}
& 76.5\std{2.3}
& 62.6\std{13.3}
& 52.5\std{5.0}
& 22.4\std{1.7}
& 47.4\std{2.0}
& 30.1\std{6.9}
& 19.7\std{1.5}
& 57.0\std{2.3}
  & 43.6\std{1.0}
\\
 DyHead \scriptsize{O365} & 5 & Full
& 43.5\std{1.0}
& 25.3\std{1.8}
& 35.8\std{0.5}
& 63.0\std{1.0}
& 56.2\std{3.9}
& 76.8\std{5.9}
& 62.5\std{8.7}
& 46.6\std{3.1}
& 28.8\std{2.2}
& 51.2\std{2.2}
& 38.7\std{4.1}
& 21.0\std{1.4}
& 53.4\std{5.2}
  & 46.4\std{1.1}
\\
 DyHead \scriptsize{O365} & 10 & Full
& 46.6\std{0.3}
& 29.0\std{2.8}
& 41.7\std{1.0}
& 65.2\std{2.5}
& 62.5\std{0.8}
& 85.4\std{2.2}
& 67.9\std{4.5}
& 47.9\std{2.2}
& 28.6\std{5.0}
& 53.8\std{1.0}
& 39.2\std{4.9}
& 27.9\std{2.3}
& 64.1\std{2.6}
  & 50.8\std{1.3}
\\
 DyHead \scriptsize{O365} & All & Full
& 53.3 
& 28.4 
& 49.5 
& 73.5 
& 77.9 
& 84.0 
& 69.2 
& 56.2 
& 43.6 
& 59.2 
& 68.9 
& 53.7 
& 73.7 
  & 60.8
\\

 \midrule \midrule
 GLIP-T & 1 & Linear
& 57.1\std{0.0}
& 15.0\std{0.3}
& 21.2\std{0.3}
& 68.3\std{1.6}
& 59.5\std{0.1}
& 72.7\std{0.3}
& 72.3\std{0.0} 
& 65.2\std{0.2}
& 26.5\std{0.1}
& 57.6\std{0.1}
& 54.1\std{0.4}
& 18.2\std{0.1}
& 47.3\std{0.2}
  & 48.9\std{0.1}
 \\
 
 GLIP-T & 3 & Linear
& 58.9\std{0.1}
& 15.3\std{0.1}
& 26.0\std{0.3}
& 70.1\std{0.5}
& 61.6\std{0.4}
& 74.7\std{0.1}
& 72.3\std{0.0} 
& 64.6\std{0.2}
& 25.9\std{0.0}
& 60.1\std{0.1}
& 51.0\std{0.2}
& 20.9\std{0.1}
& 55.5\std{0.2}
  & 50.5\std{0.1}
\\ 
 
 GLIP-T & 5 & Linear
& 59.0\std{0.1}
& 15.5\std{0.4}
& 27.6\std{0.9}
& 69.7\std{0.8}
& 61.8\std{0.1}
& 75.1\std{0.4}
& 72.3\std{0.0} 
& 62.8\std{0.5}
& 25.4\std{0.4}
& 62.5\std{0.6}
& 51.4\std{0.3}
& 19.6\std{0.6}
& 52.7\std{1.2}
  & 50.4\std{0.2}
\\ 
 
 GLIP-T & 10 & Linear
& 60.1\std{0.1}
& 14.1\std{0.1}
& 29.6\std{0.8}
& 69.5\std{0.3}
& 62.4\std{0.2}
& 76.8\std{0.1}
& 72.3\std{0.0} 
& 61.1\std{0.3}
& 25.8\std{0.2}
& 63.4\std{0.6}
& 51.0\std{0.1}
& 23.3\std{0.3}
& 55.8\std{1.3}
  & 51.2\std{0.1}
\\ 
 
 GLIP-T & All & Linear
& 65.5 
& 14.1 
& 36.5 
& 68.2 
& 67.2 
& 76.6 
& 70.2 
& 63.8 
& 29.1 
& 65.5 
& 63.5 
& 29.9 
& 66.5 
  & 55.1
\\ 
 \midrule
 GLIP-T & 1 & Prompt
& 54.4\std{0.9}
& 15.2\std{1.4}
& 32.5\std{1.0}
& 68.0\std{3.2}
& 60.0\std{0.7}
& 75.8\std{1.2}
& 72.3\std{0.0} 
& 54.5\std{3.9}
& 24.1\std{3.0}
& 59.2\std{0.9}
& 57.4\std{0.6}
& 18.9\std{1.8}
& 56.9\std{2.7}
  & 49.9\std{0.6}
\\ 
 
 GLIP-T & 3 & Prompt
& 56.8\std{0.8}
& 18.9\std{3.6}
& 37.6\std{1.6}
& 72.4\std{0.5}
& 62.8\std{1.3}
& 85.4\std{2.8}
& 64.5\std{4.6}
& 69.1\std{1.8}
& 22.0\std{0.9}
& 62.7\std{1.1}
& 56.1\std{0.6}
& 25.9\std{0.7}
& 63.8\std{4.8}
  & 53.7\std{1.3}
\\ 
 
 GLIP-T & 5 & Prompt
& 58.5\std{0.5}
& 18.2\std{0.1}
& 41.0\std{1.2}
& 71.8\std{2.4}
& 65.7\std{0.7}
& 87.5\std{2.2}
& 72.3\std{0.0} 
& 60.6\std{2.2}
& 31.4\std{4.2}
& 61.0\std{1.8}
& 54.4\std{0.6}
& 32.6\std{1.4}
& 66.3\std{2.8}
  & 55.5\std{0.5}
\\ 
 
 GLIP-T & 10 & Prompt
& 59.7\std{0.7}
& 19.8\std{1.6}
& 44.8\std{0.9}
& 72.1\std{2.0}
& 65.9\std{0.6}
& 87.4\std{1.1}
& 72.3\std{0.0} 
& 57.5\std{1.2}
& 30.0\std{1.4}
& 62.1\std{1.4}
& 57.8\std{0.9}
& 33.5\std{0.1}
& 73.1\std{1.4}
  & 56.6\std{0.2}
\\ 
 
 GLIP-T & All & Prompt
& 66.4 
& 27.6 
& 50.9 
& 70.6 
& 73.3 
& 88.1 
& 67.7 
& 64.0 
& 40.3 
& 65.4 
& 68.3 
& 50.7 
& 78.5 
  & 62.4
\\
\midrule
 GLIP-T & 1 & Full
 & 54.8\std{2.0}
& 18.4\std{1.0}
& 33.8\std{1.1}
& 70.1\std{2.9}
& 64.2\std{1.8}
& 83.7\std{3.0}
& 70.8\std{2.1}
& 56.2\std{1.8}
& 22.9\std{0.2}
& 56.6\std{0.5}
& 59.9\std{0.4}
& 18.9\std{1.3}
& 54.5\std{2.7}
  & 51.1\std{0.1}
 
 \\

 GLIP-T & 3 & Full
 & 58.1\std{0.5}
& 22.9\std{1.3}
& 40.8\std{0.9}
& 65.7\std{1.6}
& 66.0\std{0.2}
& 84.7\std{0.5}
& 65.7\std{2.8}
& 62.6\std{1.4}
& 27.2\std{2.7}
& 61.9\std{1.8}
& 60.7\std{0.2}
& 27.1\std{1.2}
& 70.4\std{2.5}
  & 54.9\std{0.2}
 
 \\
 GLIP-T & 5 & Full
 & 59.5\std{0.4}
& 23.8\std{0.9}
& 43.6\std{1.4}
& 68.7\std{1.3}
& 66.1\std{0.6}
& 85.4\std{0.4}
& 72.3\std{0.0} 
& 62.1\std{2.0}
& 27.3\std{1.2}
& 61.0\std{1.8}
& 62.7\std{1.6}
& 34.5\std{0.5}
& 66.6\std{2.3}
  & 56.4\std{0.4}
 
 \\
 GLIP-T & 10 & Full
 & 59.1\std{1.3}
& 26.3\std{1.1}
& 46.3\std{1.6}
& 67.3\std{1.5}
& 67.1\std{0.7}
& 87.8\std{0.5}
& 72.3\std{0.0} 
& 57.7\std{1.7}
& 34.6\std{1.7}
& 65.4\std{1.4}
& 61.6\std{1.0}
& 39.3\std{1.0}
& 74.7\std{2.3}
  & 58.4\std{0.2}
 
 \\
 GLIP-T & All & Full
 & 62.3 
& 31.2 
& 52.5 
& 70.8 
& 78.7 
& 88.1 
& 75.6 
& 61.4 
& 51.4 
& 65.3 
& 71.2 
& 58.7 
& 76.7 
  & 64.9
 \\
 \midrule
 \midrule

 GLIP-L & 1 & Linear
 & 63.7\std{0.1}
& 7.6\std{0.3}
& 28.1\std{0.2}
& 74.6\std{0.0}
& 60.3\std{0.0}
& 41.3\std{3.1}
& 70.2\std{1.3}
& 67.0\std{1.0}
& 71.0\std{0.0}
& 60.5\std{0.3}
& 67.9\std{0.1}
& 24.8\std{0.0}
& 66.1\std{0.0}
  & 54.1\std{0.3}
 \\
 
 GLIP-L & 3 & Linear
 & 64.8\std{0.1}
& 8.5\std{0.1}
& 33.7\std{0.2}
& 74.3\std{0.2}
& 64.1\std{0.2}
& 37.0\std{0.2}
& 69.3\std{0.0}
& 66.6\std{1.9}
& 71.2\std{0.3}
& 63.2\std{0.3}
& 68.0\std{0.1}
& 24.8\std{0.0}
& 65.9\std{0.4}
  & 54.7\std{0.2}
 
 \\ 
 GLIP-L & 5 & Linear
 & 65.0\std{0.1}
& 8.8\std{0.1}
& 33.4\std{0.3}
& 74.1\std{0.1}
& 63.8\std{0.0}
& 37.2\std{0.0}
& 69.3\std{0.0} 
& 69.2\std{0.6}
& 71.5\std{0.1}
& 64.2\std{0.3}
& 68.0\std{0.1}
& 25.3\std{0.2}
& 65.2\std{0.5}
  & 55.0\std{0.0}
 
 \\ 
 GLIP-L & 10 & Linear
 & 65.2\std{0.3}
& 11.5\std{2.3}
& 35.1\std{0.4}
& 74.0\std{0.0} 
& 64.7\std{0.0}
& 38.0\std{1.0}
& 71.7\std{1.7}
& 66.7\std{0.3}
& 72.5\std{0.3}
& 65.6\std{1.1}
& 67.9\std{0.0}
& 25.8\std{0.2}
& 67.2\std{0.3}
  & 55.8\std{0.4}
 
 \\ 
 GLIP-L & All & Linear
 & 70.9 
& 9.6 
& 42.3 
& 75.3 
& 70.5 
& 39.4 
& 69.3 
& 71.6 
& 73.9 
& 69.7 
& 72.1 
& 33.2 
& 72.3 
  & 59.2
 
 \\ 
\midrule
 GLIP-L & 1 & Prompt
 & 62.8\std{0.4}
& 18.0\std{1.8}
& 37.4\std{0.3}
& 71.9\std{2.4}
& 68.9\std{0.1}
& 81.8\std{3.4}
& 65.0\std{2.8}
& 63.9\std{0.4}
& 70.2\std{1.2}
& 67.0\std{0.4}
& 69.3\std{0.1}
& 27.6\std{0.4}
& 69.8\std{0.6}
  & 59.5\std{0.4}
 
 \\ 
 GLIP-L & 3 & Prompt
 & 65.0\std{0.5}
& 21.4\std{1.0}
& 43.6\std{1.1}
& 72.9\std{0.7}
& 70.4\std{0.1}
& 91.4\std{0.7}
& 57.7\std{3.7}
& 70.7\std{1.1}
& 69.7\std{0.9}
& 62.6\std{0.8}
& 67.7\std{0.4}
& 36.2\std{1.1}
& 68.8\std{1.5}
  & 61.4\std{0.3}
 
 \\ 
 GLIP-L & 5 & Prompt
 
 & 65.6\std{0.3}
& 19.9\std{1.6}
& 47.7\std{0.7}
& 73.7\std{0.7}
& 70.6\std{0.3}
& 86.8\std{0.5}
& 64.6\std{0.7}
& 69.4\std{3.3}
& 68.0\std{1.3}
& 67.8\std{1.5}
& 68.3\std{0.3}
& 36.6\std{1.6}
& 71.9\std{0.6}
  & 62.4\std{0.5}
 \\ 
 GLIP-L & 10 & Prompt
 & 65.9\std{0.2}
& 23.4\std{2.6}
& 50.3\std{0.7}
& 73.6\std{0.7}
& 71.8\std{0.3}
& 86.5\std{0.3}
& 70.5\std{1.1}
& 69.0\std{0.5}
& 69.4\std{2.4}
& 70.8\std{1.2}
& 68.8\std{0.6}
& 39.3\std{0.9}
& 74.9\std{2.1}
  & 64.2\std{0.4}
 
 \\ 
 GLIP-L & All & Prompt
 
 & 72.9 
& 23.0 
& 51.8 
& 72.0 
& 75.8 
& 88.1 
& 75.2 
& 69.5 
& 73.6 
& 72.1 
& 73.7 
& 53.5 
& 81.4 
  & 67.9\std{0.0}

 \\ 
\midrule
 GLIP-L & 1 & Full
 & 64.8\std{0.6}
& 18.7\std{0.6}
& 39.5\std{1.2}
& 70.0\std{1.5}
& 70.5\std{0.2}
& 69.8\std{18.0}
& 70.6\std{4.0}
& 68.4\std{1.2}
& 71.0\std{1.3}
& 65.4\std{1.1}
& 68.1\std{0.2}
& 28.9\std{2.9}
& 72.9\std{4.7}
  & 59.9\std{1.4}
 
 \\ 
 GLIP-L & 3 & Full
 
 & 65.6\std{0.6}
& 22.3\std{1.1}
& 45.2\std{0.4}
& 72.3\std{1.4}
& 70.4\std{0.4}
& 81.6\std{13.3}
& 71.8\std{0.3}
& 65.3\std{1.6}
& 67.6\std{1.0}
& 66.7\std{0.9}
& 68.1\std{0.3}
& 37.0\std{1.9}
& 73.1\std{3.3}
  & 62.1\std{0.7}
 \\ 
 GLIP-L & 5 & Full
 
 & 66.6\std{0.4}
& 26.4\std{2.5}
& 49.5\std{1.1}
& 70.7\std{0.2}
& 71.9\std{0.2}
& 88.1\std{0.0} 
& 71.1\std{0.6}
& 68.8\std{1.2}
& 68.5\std{1.7}
& 70.0\std{0.9}
& 68.3\std{0.5}
& 39.9\std{1.4}
& 75.2\std{2.7}
  & 64.2\std{0.3}

 \\ 
 GLIP-L & 10 & Full
 & 66.4\std{0.7}
& 32.0\std{1.4}
& 52.3\std{1.1}
& 70.6\std{0.7}
& 72.4\std{0.3}
& 88.1\std{0.0}
& 67.1\std{3.6}
& 64.7\std{3.1}
& 69.4\std{1.4}
& 71.5\std{0.8}
& 68.4\std{0.7}
& 44.3\std{0.6}
& 76.3\std{1.1}
  & 64.9\std{0.7}
 
 \\
 GLIP-L & All & Full
 & 69.6 
& 32.6 
& 56.6 
& 76.4 
& 79.4 
& 88.1 
& 67.1 
& 69.4 
& 65.8 
& 71.6 
& 75.7 
& 60.3 
& 83.1 
  & 68.9
 
 \\

 \bottomrule
\end{tabular}
}
\end{center}
\end{table*}

\begin{table*}[ht]
\caption{Per-dataset performance of  GLIP-T (A, B, and C). For PascalVOC, we report the mAP (IoU=0.50:0.95) using the COCO evaluation script, to be consistent with other 12 datasets. ``Linear'' denotes linear probing. ``Prompt'' denotes prompt tuning. ``Full'' denotes full-model tuning.}
\label{table:perdataset_all_2}
\begin{center}
\resizebox{\linewidth}{!}{
\begin{tabular}{l@{\hskip9pt} 
c@{\hskip9pt}c@{\hskip9pt}|l@{\hskip9pt} 
l@{\hskip9pt}l@{\hskip9pt}l@{\hskip9pt}
l@{\hskip9pt}l@{\hskip9pt}l@{\hskip9pt}l@{\hskip9pt}
l@{\hskip9pt}l@{\hskip9pt}l@{\hskip9pt}l@{\hskip9pt}l@{\hskip9pt}l@{\hskip9pt}l}
\toprule

Model & Shot & Tune & \small{PascalVOC} &
\small{AerialDrone} & 
\small{Aquarium} &
\small{Rabbits} &
\small{EgoHands} &
\small{Mushrooms} &
\small{Packages} &
\small{Raccoon} &
\small{Shellfish} &
\small{Vehicles} &
\small{Pistols} &
\small{Pothole} &
\small{Thermal} & 
Avg
\\
\midrule
 GLIP-T (A) & 1 & Linear
& 52.9\std{0.1}
& 13.2\std{0.3}
& 21.3\std{3.2}
& 65.0\std{2.0}
& 23.1\std{0.3}
& 11.4\std{0.1}
& 57.3\std{4.6}
& 53.5\std{0.7}
& 16.8\std{0.0}
& 54.1\std{0.1}
& 34.5\std{0.2}
& 5.8\std{0.1}
& 40.8\std{0.4}
  & 34.6\std{0.6}
\\

 GLIP-T (A) & 3 & Linear
& 54.6\std{0.2}
& 13.4\std{0.1}
& 28.3\std{0.1}
& 65.4\std{1.0}
& 26.0\std{0.3}
& 11.4\std{0.0}
& 50.8\std{0.7}
& 58.8\std{0.3}
& 15.8\std{0.7}
& 56.1\std{1.0}
& 34.4\std{0.9}
& 6.5\std{0.0}
& 45.8\std{0.3}
  & 35.9\std{0.2}
\\

 GLIP-T (A) & 5 & Linear
& 55.3\std{0.1}
& 14.0\std{0.3}
& 28.5\std{0.1}
& 65.2\std{1.3}
& 28.4\std{0.2}
& 11.7\std{0.0}
& 63.9\std{0.0} 
& 59.2\std{0.8}
& 16.9\std{0.2}
& 56.6\std{0.2}
& 36.9\std{0.5}
& 9.3\std{0.0}
& 43.2\std{0.3}
  & 37.6\std{0.1}
\\

 GLIP-T (A) & 10 & Linear
& 56.8\std{0.2}
& 14.3\std{0.2}
& 29.0\std{0.1}
& 67.0\std{0.1}
& 29.2\std{0.1}
& 11.6\std{0.1}
& 64.5\std{0.3}
& 59.7\std{0.7}
& 16.6\std{0.7}
& 56.9\std{0.0}
& 33.2\std{1.5}
& 7.4\std{0.1}
& 46.2\std{0.8}
  & 37.9\std{0.2}
\\

 GLIP-T (A) & All & Linear
& 62.0 
& 15.1 
& 32.2 
& 66.1 
& 40.9 
& 12.1 
& 66.9 
& 60.5 
& 22.5 
& 62.4 
& 49.8 
& 17.1 
& 65.7 
  & 44.1\std{0.0}
\\
\midrule
 GLIP-T (A) & 1 & Prompt
& 52.1\std{0.5}
& 11.4\std{0.2}
& 23.7\std{0.6}
& 66.6\std{0.2}
& 21.0\std{0.2}
& 8.6\std{0.6}
& 46.7\std{0.1}
& 53.2\std{0.2}
& 17.1\std{0.7}
& 58.8\std{0.2}
& 37.9\std{0.3}
& 6.0\std{0.2}
& 38.3\std{0.4}
  & 34.0\std{0.1}
\\

 GLIP-T (A) & 3 & Prompt
& 54.9\std{0.1}
& 13.4\std{2.5}
& 25.9\std{0.2}
& 65.9\std{0.5}
& 22.7\std{0.1}
& 33.6\std{1.4}
& 46.6\std{0.0}
& 53.7\std{0.4}
& 18.5\std{0.8}
& 58.2\std{0.6}
& 38.1\std{0.5}
& 6.2\std{0.1}
& 42.4\std{0.2}
  & 36.9\std{0.5}
\\

 GLIP-T (A) & 5 & Prompt
& 55.6\std{0.2}
& 13.6\std{0.4}
& 26.1\std{0.4}
& 65.7\std{1.5}
& 24.5\std{0.4}
& 56.9\std{2.6}
& 60.5\std{0.6}
& 55.2\std{0.2}
& 19.0\std{1.5}
& 57.0\std{0.8}
& 36.4\std{1.4}
& 6.3\std{0.1}
& 43.2\std{0.1}
  & 40.0\std{0.3}
\\

 GLIP-T (A) & 10 & Prompt
& 56.6\std{0.1}
& 15.8\std{0.8}
& 26.2\std{0.1}
& 68.0\std{0.6}
& 24.4\std{0.1}
& 41.2\std{12.5}
& 60.3\std{0.9}
& 55.9\std{0.4}
& 19.6\std{1.6}
& 57.5\std{1.0}
& 36.1\std{0.3}
& 6.0\std{0.1}
& 42.4\std{1.2}
  & 39.2\std{0.9}
\\

 GLIP-T (A) & All & Prompt
& 58.8 
& 16.4 
& 28.7 
& 69.5 
& 28.8 
& 56.9 
& 60.9 
& 56.3 
& 20.5 
& 60.7 
& 43.3 
& 10.4 
& 51.2 
  & 43.3
\\
\midrule
 GLIP-T (A) & 1 & Full
& 44.8\std{0.7}
& 16.9\std{1.2}
& 28.0\std{1.0}
& 64.6\std{1.6}
& 54.1\std{1.5}
& 64.1\std{12.0}
& 55.8\std{0.6}
& 55.6\std{1.8}
& 21.6\std{0.9}
& 53.4\std{1.3}
& 43.8\std{0.9}
& 10.9\std{1.2}
& 52.3\std{4.7}
  & 43.5\std{1.2}
\\

 GLIP-T (A) & 3 & Full
& 49.5\std{0.6}
& 23.3\std{1.4}
& 36.7\std{1.2}
& 62.5\std{1.6}
& 59.9\std{1.1}
& 84.1\std{1.3}
& 60.2\std{1.1}
& 45.0\std{2.6}
& 26.5\std{1.9}
& 54.4\std{0.7}
& 44.6\std{3.7}
& 23.6\std{0.7}
& 63.5\std{2.7}
  & 48.8\std{0.3}
\\

 GLIP-T (A) & 5 & Full
& 50.8\std{0.5}
& 25.3\std{0.7}
& 41.2\std{0.8}
& 62.4\std{0.9}
& 60.4\std{0.9}
& 86.4\std{2.3}
& 59.2\std{8.5}
& 44.7\std{2.5}
& 28.2\std{0.7}
& 55.6\std{2.0}
& 51.7\std{0.8}
& 27.0\std{0.8}
& 62.1\std{6.0}
  & 50.4\std{0.6}
\\

 GLIP-T (A) & 10 & Full
& 51.7\std{0.3}
& 29.9\std{2.4}
& 44.3\std{0.8}
& 67.8\std{2.7}
& 64.1\std{0.3}
& 87.9\std{0.3}
& 71.3\std{2.0}
& 47.0\std{4.2}
& 28.8\std{2.0}
& 56.9\std{0.9}
& 52.3\std{0.4}
& 29.1\std{2.9}
& 72.7\std{2.2}
  & 54.1\std{0.4}
\\

 GLIP-T (A) & All & Full
& 55.1 
& 35.3 
& 50.9 
& 78.0 
& 78.0 
& 86.3 
& 75.2 
& 54.8 
& 44.1 
& 61.4 
& 69.3 
& 57.3 
& 80.6 
  & 63.6
\\
\midrule
\midrule

 GLIP-T (B) & 1 & Linear
& 54.0\std{0.1}
& 6.6\std{0.0}
& 17.2\std{0.0}
& 73.3\std{0.7}
& 23.7\std{0.7}
& 63.6\std{0.2}
& 51.5\std{0.0}
& 51.8\std{0.2}
& 25.5\std{0.1}
& 56.4\std{0.1}
& 45.2\std{1.0}
& 6.7\std{0.1}
& 56.5\std{0.4}
  & 40.9\std{0.2}
\\

 GLIP-T (B) & 3 & Linear
& 54.9\std{0.0}
& 6.6\std{0.0}
& 25.2\std{0.2}
& 73.1\std{0.3}
& 29.3\std{0.4}
& 63.3\std{0.1}
& 55.3\std{3.6}
& 56.1\std{0.4}
& 24.8\std{0.4}
& 57.5\std{0.6}
& 44.8\std{0.1}
& 6.9\std{0.2}
& 58.5\std{0.3}
  & 42.8\std{0.3}
\\

 GLIP-T (B) & 5 & Linear
& 56.0\std{0.5}
& 6.6\std{0.0}
& 25.7\std{0.3}
& 72.9\std{0.8}
& 28.4\std{0.1}
& 62.7\std{0.2}
& 70.5\std{1.2}
& 56.1\std{0.3}
& 25.4\std{0.5}
& 58.6\std{0.2}
& 46.8\std{0.5}
& 9.4\std{0.9}
& 52.8\std{0.4}
  & 44.0\std{0.2}
\\

 GLIP-T (B) & 10 & Linear
& 57.3\std{0.2}
& 6.6\std{0.0}
& 27.8\std{0.9}
& 75.8\std{0.5}
& 30.1\std{0.2}
& 62.8\std{0.4}
& 67.8\std{1.3}
& 53.2\std{0.2}
& 24.0\std{0.1}
& 61.5\std{1.4}
& 43.9\std{0.3}
& 7.6\std{0.1}
& 58.4\std{0.5}
  & 44.4\std{0.3}
\\

 GLIP-T (B) & All & Linear
& 64.3 
& 6.6 
& 35.6 
& 73.9 
& 44.9 
& 62.8 
& 73.6 
& 63.9 
& 34.2 
& 65.0 
& 61.8 
& 20.5 
& 66.6 
  & 51.8
\\

\midrule

 GLIP-T (B) & 1 & Prompt
& 52.7\std{0.4}
& 16.1\std{0.8}
& 25.2\std{0.3}
& 72.5\std{0.4}
& 56.4\std{0.5}
& 74.5\std{1.0}
& 56.2\std{4.5}
& 56.5\std{1.3}
& 22.3\std{1.5}
& 55.0\std{0.8}
& 53.0\std{1.3}
& 7.1\std{0.5}
& 54.9\std{0.8}
  & 46.4\std{0.4}
\\

 GLIP-T (B) & 3 & Prompt
& 54.7\std{0.9}
& 16.6\std{0.6}
& 33.8\std{0.3}
& 76.7\std{1.0}
& 55.9\std{0.6}
& 77.2\std{4.2}
& 59.5\std{5.6}
& 55.7\std{2.7}
& 24.2\std{1.2}
& 56.9\std{0.7}
& 51.3\std{1.4}
& 18.4\std{0.6}
& 56.6\std{1.7}
  & 49.0\std{0.7}
\\

 GLIP-T (B) & 5 & Prompt
& 57.4\std{0.3}
& 20.0\std{1.5}
& 35.9\std{1.3}
& 76.0\std{0.4}
& 58.2\std{0.8}
& 78.7\std{4.2}
& 61.4\std{1.2}
& 56.5\std{1.5}
& 27.2\std{0.8}
& 55.0\std{4.7}
& 53.6\std{1.8}
& 21.4\std{0.3}
& 56.4\std{1.0}
  & 50.6\std{0.4}
\\

 GLIP-T (B) & 10 & Prompt
& 57.8\std{0.6}
& 22.5\std{0.7}
& 39.1\std{0.8}
& 74.7\std{1.3}
& 58.8\std{0.8}
& 85.6\std{1.3}
& 59.6\std{0.0}
& 56.7\std{1.5}
& 32.4\std{0.8}
& 59.3\std{1.8}
& 52.4\std{0.5}
& 20.7\std{1.0}
& 66.1\std{1.8}
  & 52.8\std{0.1}
\\

 GLIP-T (B) & All & Prompt
& 64.6 
& 18.2 
& 47.3 
& 71.3 
& 70.1 
& 85.6 
& 59.6 
& 65.0 
& 37.9 
& 61.3 
& 64.6 
& 39.0 
& 76.4 
  & 58.5
\\
\midrule

 GLIP-T (B) & 1 & Full
& 48.4\std{1.9}
& 16.6\std{0.6}
& 31.8\std{1.7}
& 70.9\std{1.4}
& 55.3\std{0.4}
& 78.8\std{2.7}
& 66.3\std{1.6}
& 48.1\std{6.9}
& 23.3\std{1.3}
& 57.0\std{0.8}
& 52.9\std{0.6}
& 12.9\std{0.4}
& 61.0\std{1.8}
  & 48.0\std{0.7}
\\

 GLIP-T (B) & 3 & Full
& 51.7\std{0.8}
& 23.4\std{2.6}
& 37.2\std{1.0}
& 69.5\std{1.0}
& 59.6\std{0.7}
& 85.4\std{0.4}
& 62.4\std{1.1}
& 56.5\std{1.5}
& 30.0\std{1.0}
& 57.6\std{1.2}
& 54.7\std{1.6}
& 24.5\std{1.3}
& 64.3\std{1.8}
  & 52.1\std{0.4}
\\

 GLIP-T (B) & 5 & Full
& 52.9\std{0.7}
& 27.4\std{0.7}
& 41.5\std{0.6}
& 68.4\std{1.4}
& 61.9\std{0.5}
& 81.0\std{3.3}
& 69.3\std{3.5}
& 61.2\std{2.6}
& 26.9\std{1.9}
& 58.1\std{0.3}
& 57.4\std{1.7}
& 28.3\std{1.5}
& 57.3\std{2.4}
  & 53.2\std{0.7}
\\

 GLIP-T (B) & 10 & Full
& 53.9\std{1.1}
& 28.2\std{1.3}
& 43.1\std{0.8}
& 69.0\std{2.1}
& 65.4\std{1.4}
& 87.3\std{0.6}
& 65.1\std{2.1}
& 52.3\std{3.2}
& 30.6\std{0.7}
& 60.2\std{2.2}
& 53.0\std{2.5}
& 34.2\std{1.9}
& 71.8\std{2.3}
  & 54.9\std{0.6}
\\

 GLIP-T (B) & All & Full
& 56.9 
& 28.7 
& 54.0 
& 68.3 
& 78.4 
& 88.1 
& 72.7 
& 57.7 
& 41.2 
& 63.8 
& 69.0 
& 59.8 
& 75.8 
  & 62.7
\\

\midrule\midrule
 GLIP-T (C) & 1 & Linear
& 57.0\std{0.2}
& 6.4\std{0.1}
& 21.1\std{0.4}
& 74.2\std{0.0}
& 60.9\std{0.1}
& 24.6\std{0.1}
& 64.0\std{0.0}
& 52.0\std{0.1}
& 21.2\std{0.1}
& 55.6\std{0.2}
& 50.9\std{0.3}
& 14.6\std{0.0}
& 68.7\std{0.6}
  & 43.9\std{0.1}
\\

 GLIP-T (C) & 3 & Linear
& 59.0\std{0.1}
& 8.2\std{0.4}
& 28.4\std{0.2}
& 74.2\std{0.0}
& 61.5\std{0.1}
& 24.2\std{0.3}
& 64.0\std{0.0}
& 57.8\std{0.6}
& 20.9\std{0.1}
& 57.1\std{0.5}
& 49.3\std{0.2}
& 15.7\std{0.1}
& 69.5\std{0.5}
  & 45.4\std{0.0}
\\

 GLIP-T (C) & 5 & Linear
& 59.6\std{0.0}
& 6.5\std{0.1}
& 29.9\std{0.6}
& 74.1\std{1.5}
& 61.9\std{0.0}
& 24.9\std{0.1}
& 64.9\std{1.3}
& 52.0\std{0.3}
& 21.7\std{0.4}
& 63.4\std{0.3}
& 48.5\std{1.2}
& 22.2\std{0.3}
& 67.6\std{0.7}
  & 45.9\std{0.1}
\\

 GLIP-T (C) & 10 & Linear
& 60.8\std{0.2}
& 7.6\std{0.5}
& 31.6\std{0.1}
& 74.3\std{1.2}
& 63.2\std{0.1}
& 25.3\std{0.2}
& 65.8\std{0.6}
& 58.2\std{2.8}
& 22.6\std{0.3}
& 62.6\std{0.3}
& 46.0\std{0.1}
& 20.0\std{0.4}
& 69.4\std{1.1}
  & 46.7\std{0.2}
\\
 GLIP-T (C) & All & Linear
& 66.4 
& 8.2 
& 38.2 
& 71.0 
& 68.5 
& 37.7 
& 64.0 
& 59.7 
& 32.5 
& 66.1 
& 62.4 
& 32.4 
& 78.2 
  & 52.7
 \\
 \midrule
 GLIP-T (C) & 1 & Prompt
& 52.6\std{1.0}
& 13.3\std{0.8}
& 30.8\std{1.5}
& 70.4\std{0.9}
& 60.3\std{0.4}
& 74.5\std{3.1}
& 71.1\std{1.4}
& 58.8\std{0.2}
& 24.8\std{1.4}
& 58.4\std{1.1}
& 51.8\std{1.5}
& 22.8\std{1.1}
& 68.2\std{0.1}
  & 50.6\std{0.4}
\\

 GLIP-T (C) & 3 & Prompt
& 57.4\std{0.2}
& 18.9\std{1.3}
& 36.2\std{1.2}
& 74.0\std{2.6}
& 64.0\std{1.1}
& 84.6\std{0.7}
& 64.1\std{3.7}
& 59.2\std{3.8}
& 23.0\std{2.6}
& 61.2\std{1.4}
& 53.1\std{1.7}
& 27.0\std{1.1}
& 65.5\std{1.4}
  & 52.9\std{0.4}
\\

 GLIP-T (C) & 5 & Prompt
& 58.8\std{1.0}
& 20.2\std{0.7}
& 41.3\std{1.3}
& 73.2\std{1.4}
& 64.6\std{1.8}
& 82.3\std{2.7}
& 69.1\std{5.1}
& 58.0\std{2.7}
& 27.2\std{4.0}
& 59.2\std{1.9}
& 53.7\std{0.5}
& 26.2\std{2.3}
& 66.5\std{2.5}
  & 53.9\std{0.5}
\\

 GLIP-T (C) & 10 & Prompt
& 59.8\std{0.6}
& 21.9\std{3.1}
& 42.8\std{0.7}
& 73.1\std{0.9}
& 66.9\std{0.5}
& 85.7\std{3.6}
& 69.9\std{2.1}
& 58.5\std{1.8}
& 25.7\std{1.4}
& 61.3\std{1.1}
& 54.1\std{0.4}
& 30.1\std{3.7}
& 74.9\std{0.2}
  & 55.8\std{0.9}
\\

 GLIP-T (C) & All & Prompt
& 67.3
& 24.8
& 49.0
& 72.2
& 73.2
& 82.5
& 72.2
& 61.1
& 42.6
& 64.5
& 68.8
& 51.8
& 80.7
  & 62.4
\\

\midrule
 GLIP-T (C) & 1 & Full
& 52.5\std{0.4}
& 16.2\std{1.2}
& 34.5\std{1.3}
& 68.9\std{1.1}
& 64.2\std{1.2}
& 80.9\std{1.3}
& 65.9\std{3.9}
& 51.9\std{1.2}
& 22.3\std{3.1}
& 56.3\std{1.3}
& 55.7\std{1.2}
& 20.8\std{1.3}
& 55.0\std{4.2}
  & 49.6\std{0.2}
\\

 GLIP-T (C) & 3 & Full
& 57.1\std{0.4}
& 23.9\std{0.2}
& 39.2\std{0.1}
& 68.2\std{0.7}
& 65.9\std{0.6}
& 85.4\std{0.3}
& 68.3\std{0.2}
& 52.0\std{2.9}
& 30.8\std{1.8}
& 59.0\std{1.3}
& 54.9\std{1.1}
& 29.5\std{3.3}
& 64.8\std{3.0}
  & 53.8\std{0.1}
\\
 GLIP-T (C) & 5 & Full
& 57.6\std{0.7}
& 27.6\std{1.1}
& 43.6\std{0.3}
& 67.8\std{2.0}
& 66.4\std{0.4}
& 84.2\std{0.4}
& 67.6\std{2.6}
& 55.4\std{2.7}
& 27.1\std{5.2}
& 60.4\std{2.7}
& 59.8\std{0.8}
& 37.8\std{1.1}
& 57.0\std{6.3}
  & 54.8\std{0.8}
\\
 GLIP-T (C) & 10 & Full
& 57.1\std{0.4}
& 31.9\std{1.3}
& 47.9\std{1.0}
& 66.7\std{4.1}
& 67.7\std{0.4}
& 86.1\std{2.8}
& 63.2\std{3.4}
& 52.2\std{4.3}
& 35.5\std{1.1}
& 61.2\std{0.7}
& 58.6\std{0.9}
& 38.9\std{1.6}
& 75.8\std{3.6}
  & 57.1\std{0.9}
\\
 GLIP-T (C) & All & Full
& 62.3 
& 29.1 
& 53.8 
& 72.7 
& 78.4 
& 85.8 
& 68.6 
& 60.7 
& 43.6 
& 65.9 
& 72.2 
& 55.9 
& 81.1 
  & 63.9
\\

\bottomrule
\end{tabular}
}
\end{center}
\end{table*}

\clearpage

\end{document}